%% file: main.tex
\documentclass[runningheads]{llncs}

\usepackage{eccv}

\usepackage{eccvabbrv}

\usepackage{graphicx}
\usepackage{amsmath}
\usepackage{amssymb}
\usepackage{multirow} 
\usepackage{booktabs}
\usepackage{soul}  
\usepackage{xcolor}
\usepackage[accsupp]{axessibility}  
\usepackage{sidecap} 
\usepackage{floatrow}  

\usepackage{tabularx}

\definecolor{best}{HTML}{d7191c}
\definecolor{secondbest}{HTML}{0571b0}

\newcommand{\best}[1]{\textbf{#1}}

\newcommand{\posd}[1]{\textcolor{gray}{\scriptsize (}\textcolor{ForestGreen}{\scriptsize #1}\textcolor{gray}{\scriptsize )}}

\newcommand{\posc}[1]{\textcolor{green!70!black}{#1}}
\newcommand{\negc}[1]{\textcolor{red!85!black}{#1}}
\newcommand{\zeroc}[1]{\textcolor{black}{#1}}

\usepackage[breaklinks,colorlinks,citecolor=eccvblue]{hyperref}

\usepackage{orcidlink}
\usepackage{placeins}

\begin{document}

\title{Unsupervised Semantic Segmentation Facilitates Model Understanding}


\author{Xiaoyan Yu\inst{1,2,3,*}\orcidlink{0000-0001-8196-663X}
\and
Jannik Franzen\inst{1,2,4,5}\orcidlink{0000-0002-0761-641X}
\and
Lisa Mais\inst{1,2,5}\orcidlink{0000-0002-9281-2668}
\and
Peter Hirsch\inst{1,2}\orcidlink{0000-0002-2353-5310}
\and
Nick Lechtenbörger\inst{5}\orcidlink{0009-0006-4549-8633}
\and
Andreas Mardt\inst{1,2}\orcidlink{0000-0002-7353-6063}
\and
Dagmar Kainmueller\inst{1,2,5,*}\orcidlink{0000-0002-9830-2415}
}

\authorrunning{X.~Yu et al.}

\institute{Max-Delbrück-Center (MDC), Berlin, Germany\\
\email{\{firstname.lastname\}@mdc-berlin.de} \textsuperscript{*} Corresponding authors \and
Helmholtz Imaging \and  Humboldt-Universität zu Berlin, Berlin, Germany \and Charité Universitätsmedizin, Berlin, Germany \and University of Potsdam, Potsdam, Germany }

\maketitle

\begin{abstract}

Self-supervised learning (SSL) has produced a diverse landscape of vision transformers (ViTs) whose pretrained representations support a wide range of downstream tasks. Towards a better understanding of these models, a body of works has assessed the mechanics of their self-attention as well as which types of information they capture across their representations, revealing, e.g.,\  stark differences between models trained with contrastive learning (CL) vs.\ masked image modeling (MIM). However, the total of these advances on model understanding has to date not yet fully permeated a larger community, where, e.g., insights that are specific to CL models are still at times generalized to MIM models. To make model understanding straightforward and intuitive for a broad community, we propose a simple and easily interpretable visualization protocol.
Our protocol is based on visualizing unsupervised semantic segmentation results -- yet by no means do we focus on top segmentation performance. Instead, our protocol allows us to easily convey model behavior that consistently emerges across images. Benchmarked on a diverse set of SSL models across layers and representations, our protocol allows us to gain novel insights into distinct positional biases and scaling behaviors, including, e.g.,\ strong boundary artifacts in DINOv3-Large model tokens. These novel insights come on top of more easily conveying a range of previous findings. 
Our protocol further allows us to clearly visually convey and distinguish between \emph{positional effects} and the closely related yet distinct \emph{locality bias}, the latter being much more extensively studied in the literature so far.
%
Our protocol is publicly available\footnote{\url{https://github.com/Kainmueller-Lab/ssl-rep-seg}}, serving to catalyze further model understanding for a broad community. 


\end{abstract}

\input{content/1_intro}
\input{content/3_method}

\input{content/4_experiments}

\input{content/5_conclusion}

\FloatBarrier
\bibliographystyle{splncs04}
\bibliography{main}

\clearpage
\appendix
\input{content/7_appendix}

\end{document}

%% file: content/1_intro.tex
\section{Introduction}

Advances in self-supervised learning (SSL) have led to a diverse range of vision transformers (ViTs) that capture rich visual semantics without relying on human-annotated labels.
These pretrained models have been widely and successfully adopted across a broad spectrum of downstream vision tasks, ranging from image-level tasks such as image classification to dense pixel-level tasks such as object detection and segmentation~\cite{liu2021swin,wang2021pyramid,hahn2024boosting,wang2023cut}.
To generalize effectively across tasks, models must encode semantic information at both the global image level and the local patch level.
Here, semantic information refers to high-level, meaning-related aspects of an image, including instance-, class- and parts-level information about objects and background as well as broader scene properties such as spatial arrangement and conceptual similarity across images.

Many works have investigated how ViTs are able to achieve this.
A particular focus has been on dissecting short-range vs.\ long-range attention \cite{xie2023darksecrets,park2023whatssllearn,walmer2023teaching}, where short-range (\emph{local}) attention has been termed \emph{locality bias}.
Locality bias constitutes \emph{relative} positional information -- i.e., an encoding of what is close to a particular location \emph{within an individual image}. It has been found to be strong in MIM as opposed to CL\footnote{Note, by \emph{CL} we also refer to joint embedding objectives that do not feature an explicit push force, see e.g.\ \cite{wu2025simplifying} for a respective discussion.} (while both model types feature long-range semantically meaningful attention), serving to explain MIM's improved performance in some downstream tasks \cite{park2023whatssllearn}. 
Related yet distinct from locality bias, \emph{positional effect}, as formally defined and studied in \cite{doshi2026svd}, refers to positional information that is \emph{absolute} in the sense that it appears consistently \emph{across images}. Findings on the benefits of locality bias over positional effect have e.g.\ prompted a pivot from traditional (absolute) positional embeddings to rotary (relative) ones \cite{su2024roformer,simeoni2025dinov3}. 
While \emph{locality bias} has been conveyed in an easily interpretable form by means of attention map- and attention matrix visualizations (as e.g.,\ in \cite{bolya2025perception}), \emph{positional effect} is lacking an established distinct and intuitive visualization to date, hindering straightforward insights into  model behavior.

Besides positional information and respective attention mechanisms, the scaling behavior of SSL models has also been scrutinized, finding that large models often underperform their base variants on dense prediction tasks. This has been attributed to decreased patch consistency observed in some large models \cite{simeoni2025dinov3}; it could also be that some large models capture more fine-grained semantic structure -- a non-detrimental effect not reflected by standard benchmarks \cite{rossetti2024hierarchy}. 
However, large models' underperformance versus their base variants has not yet been studied under a hypothesis of potentially detrimental positional effects. 

Despite the manifold advances on model understanding discussed above, the community at times still attributes the widely known findings of \cite{amir2021deep} to general SSL models albeit they mainly apply to CL, e.g.\ implying that key embeddings capture strong semantic information in general  \cite{zhou2024image} while this only holds for CL, or that positional information degrades in later ViT layers in general while, again, this mainly holds for CL \cite{li2026does}.

Towards making model understanding easy and intuitive for a broad community, we propose a simple and straightforward visualization protocol. Based on unsupervised semantic segmentation by scalable k-means clustering, our protocol reveals behaviors that emerge consistently across images, thus allowing us to convey not just classical semantic information, but also positional effect. 
The protocol itself is not new -- in fact it has been standard practice in the unsupervised segmentation community~\cite{hamilton2022unsupervised,seong2023leveraging,hahn2024boosting} -- albeit with an exclusive focus on downstream segmentation performance. Instead, we borrow this protocol for the purpose of mechanistic model understanding. 
%
Previous visualization techniques for model understanding either do not establish correspondences across images (e.g.,\ \cite{caron2021emerging} as well as all kinds of attention map- and matrix visualizations), or the few that do (e.g.,\ \cite{doshi2026svd})
are not designed to facilitate comparisons across models, or are hard to interpret in this regard. 
Our protocol for the first time clearly conveys positional effect as opposed to locality bias. 

We benchmark our protocol on a range of  models, for their keys, queries, values and tokens across all layers.
Our benchmark includes eight SSL models (MAE~\cite{he2022masked}, MoCov3~\cite{chen2021empirical}, Mugs~\cite{zhou2022mugs}, iBOT~\cite{zhou2021ibot}, DINO~\cite{caron2021emerging}, DINOv2~\cite{oquab2023dinov2}, DINOv2+reg~\cite{darcet2023vision}, DINOv3~\cite{simeoni2025dinov3}) and two baselines (a supervised ViT~\cite{dosovitskiy2020image} and CLIP~\cite{ilharco2021openclip}).
Our results convey many interesting findings from related work in a coherent, easily interpretable and intuitive form, including 
(i)~optimal semantic structure typically emerging in intermediate layers rather than the final layer~\cite{bolya2025perception,simeoni2025dinov3}, 
%
(ii)~block artifacts in CLIP and DINOv2 tokens~\cite{yang2024denoising}, 
(iii)~a progressive loss of fine-grained semantic detail in deeper layers of supervised and CLIP models~\cite{walmer2023teaching}, 
(iv)~detrimental scaling behavior due to a loss of patch consistency~\cite{simeoni2025dinov3}, and 
(v)~pretext-task-specific locality biases and attention mechanisms \cite{amir2021deep,xie2023darksecrets,park2023whatssllearn,walmer2023teaching}. For the latter, we newly find that strong positional effect dominates keys and queries in MIM (cf.\ \cref{fig:protocol}), which constitutes an "upstream" cause for the respective known locality bias. Beyond conveying and refining previous findings, our results yield novel insights into detrimental scaling behavior, suggesting that strong positional effects specific to DINOv3-\emph{Large} tokens cause underperformance as compared to the respective Base model (cf.\ \cref{fig:scaling_different_models}). 
 
In summary, we contribute a simple and straightforward visualization protocol for qualitative assessment of ViT representations, complemented by contextual quantitative measures. A broad benchmark shows its power to facilitate model understanding and catalyze respective novel insights. 

%% file: content/3_method.tex
\section{Proposed protocol for model understanding}
\begin{figure}
  \centering
  \includegraphics[width=\columnwidth]{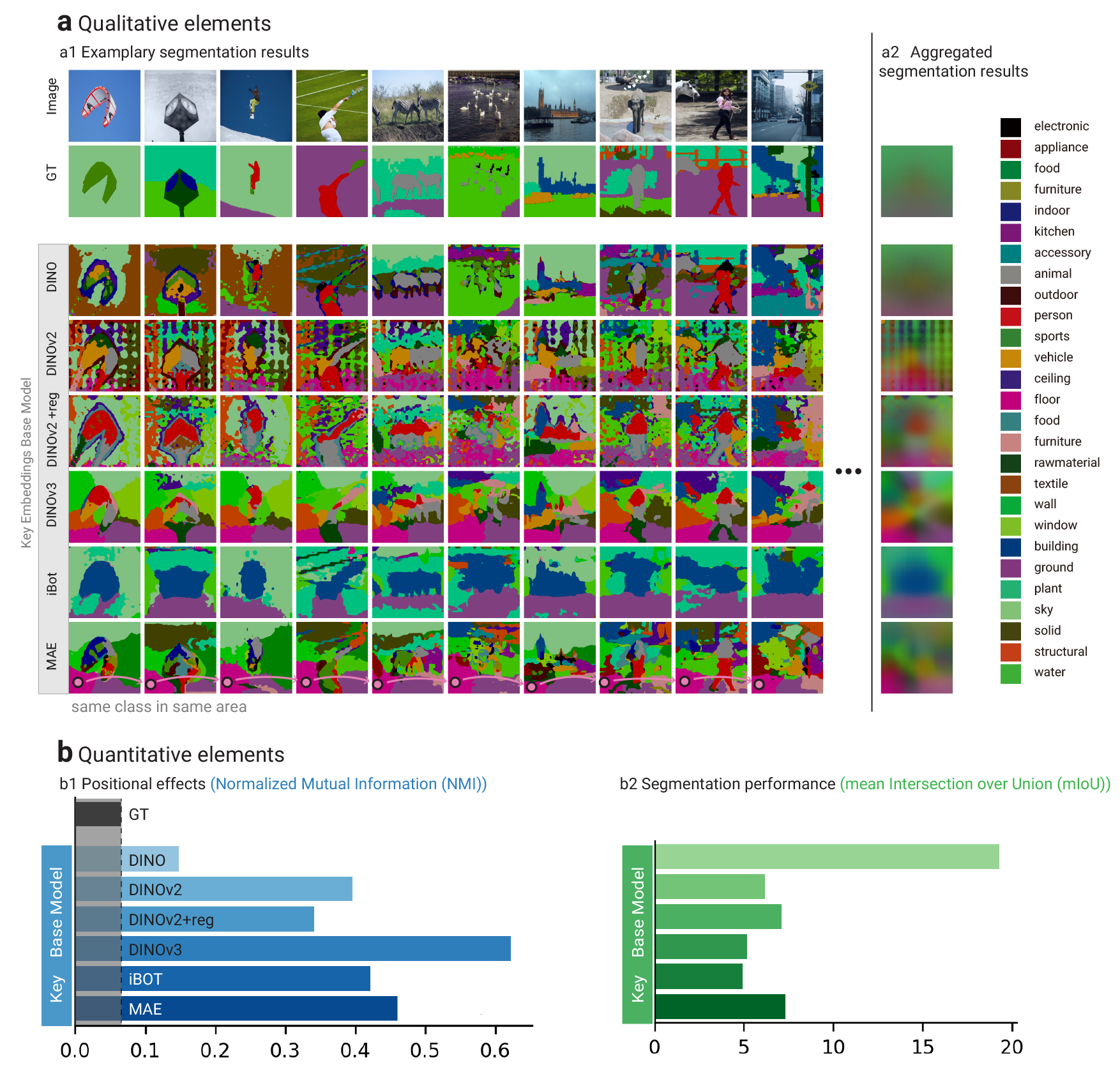}
    \caption{\textbf{Overview of our protocol.} We here introduce the individual elements of the protocol, and at the same time showcase exemplary insights it facilitates. To this end, across rows, we vary the SSL training paradigm at fixed ViT-B architecture and fixed ViT layer selection strategy by best mIoU.    
    \textbf{a)} Qualitative elements: (a1) Joint visualization of unsupervised semantic segmentation results on exemplary sets of images alongside respective ground truth label maps -- \textbf{This reveals strong positional effects in key embeddings of MIM-based models}, including DINOv2, DINOv2+reg, DINOv3, iBOT and MAE. E.g., MAE key embeddings  (bottom row) in the bottom-left corner of each individual image consistently cluster into a single class. In contrast, DINO, a CL model, exhibits substantially less positional effect in favor of more semantically meaningful structure. (a2) Aggregated visualization of unsupervised segmentation results as well as ground truth labels -- This serves to confirm that the behavior observed in a1 extends across the whole dataset.  \textbf{b)}~Quantitative elements: (b1) Aggregate positional effect -- This serves to quantitatively compare the positional effects as visually observed through the qualitative elements of the protocol, including the ground truth reference effect. (b2) Aggregate segmentation performance -- This serves to quantitatively compare semantic structure as visually conveyed through the qualitative elements. For further insights, see \cref{sec:results} and \cref{fig:pos_bias_layer_wise,fig:scaling_different_models}.
}
   \label{fig:protocol}
\end{figure}
\begin{figure}[b]
\centering
\setlength{\columnsep}{0cm}
\floatbox[{\capbeside\thisfloatsetup{capbesideposition={left,center},capbesidewidth=0.62\columnwidth,}}]{figure}
{\caption{Canonical embeddings produced by a ViT: keys, queries, and values of the Multi-Head Attention (MHA) blocks, and tokens, defined as the feed-forward network (FFN) outputs for patch tokens. Each of these embedding types is at hand at each layer of the ViT. To aggregate per-head embeddings yielded by the MHA block, for any given layer and type, we concatenate embeddings along the attention dimension. We then run cluster-based probing independently for each embedding type and at each layer of the ViT.
}\label{fig:overview}}
{
\includegraphics[width=0.36\columnwidth]{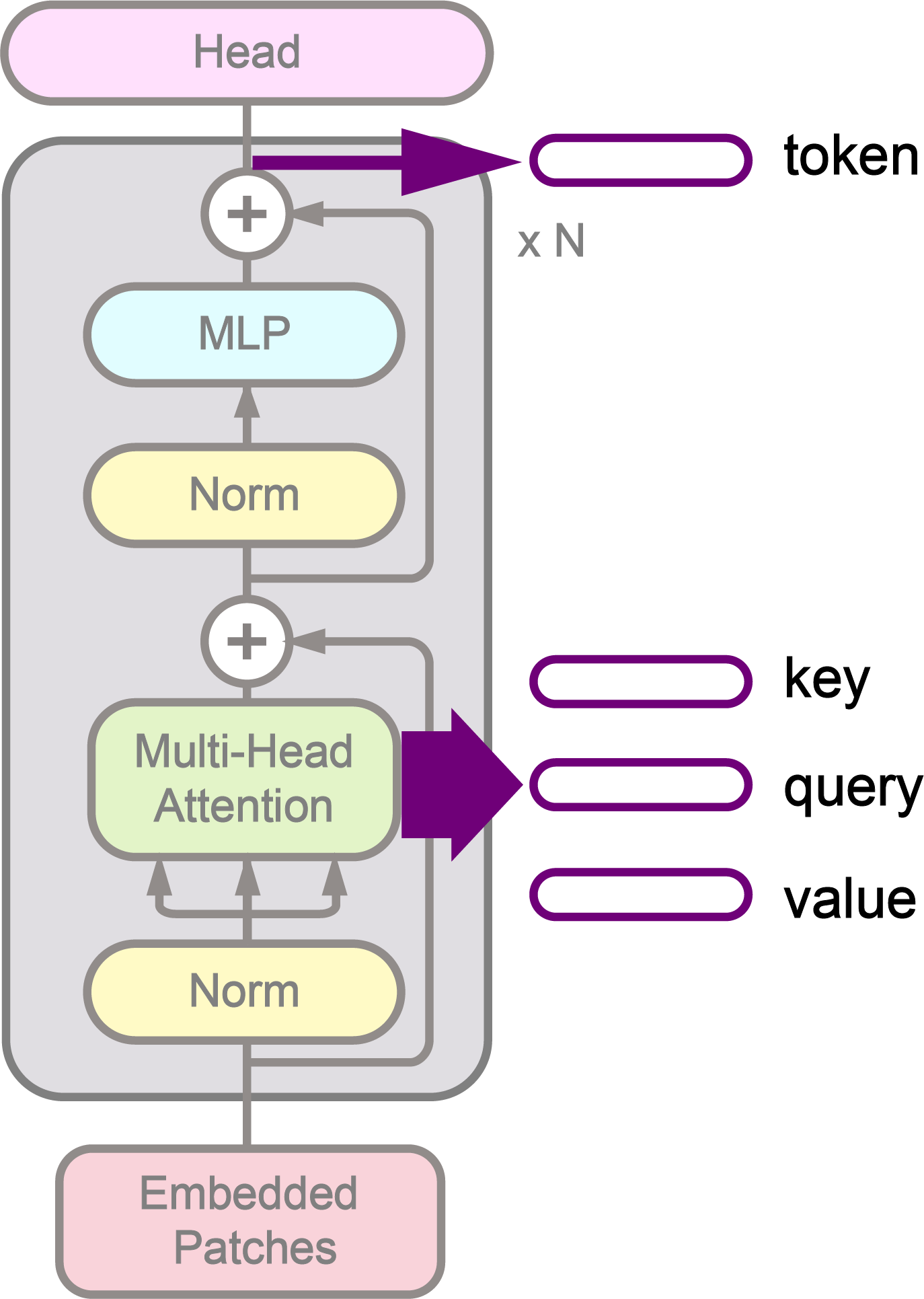}
}

\end{figure}

Our goal is to facilitate and extend our understanding of how both semantic and positional information are encoded across different components of  ViT~\cite{dosovitskiy2020image} backbones, and how various SSL paradigms influence this structure.
Our protocol to facilitate respective model understanding comprises the following core elements:
\begin{enumerate}
    \item Unsupervised semantic segmentation by across-image cluster-based probing of embeddings (cf.\ Sec.\ \ref{subsec:unsup-seg}), and
    \item Joint visualization of segmentation results on exemplary sets of images alongside ground truth labels (cf.\ Sec.\ \ref{subsec:visualization}).
\end{enumerate}
Putting visualizations of unsupervised segmentation results on a (random) set of exemplary images next to each other reveals where embeddings from same clusters localize across images, intuitively visualizing dominant positional effects while simultaneously enabling standard qualitative assessment of segmentation performance, see \cref{fig:protocol}(a1). 
To further capture a global, dataset-wide view on top of a (necessarily small) set of examples, we complement the above with: 
\begin{enumerate}
\setcounter{enumi}{2}
    \item aggregate visualization of results on the full dataset (cf.\ Sec.\ \ref{subsec:visualization}),
    \item aggregate quantitative measure of positional effect (cf.\ Sec.\ \ref{subsec:quantitative_measures}), and
    \item aggregate quantitative measure of segmentation quality (cf.\ Sec.\ \ref{subsec:quantitative_measures}), 
\end{enumerate}
see Figs.\ \ref{fig:protocol}(a2) and \ref{fig:protocol}(b).
Importantly, all elements of the protocol (except  aggregate segmentation quality) are also displayed for the ground truth labels, serving as a reference to gauge dataset-inherent spatial bias.
The protocol applies to any kind of embedding from any ViT layer (cf.\ Fig.\ \ref{fig:overview}). 
It can also be applied multiple times, to different embeddings to facilitate comparative insights (cf.\ \cref{sec:protocol-scenarios}). 

\subsection{Unsupervised semantic segmentation}
\label{subsec:unsup-seg}
We leverage unsupervised semantic segmentation as a tool to probe embedding spaces. 
We argue that any fine-tuning regime toward a specific downstream task and dataset distorts the original embedding space, hence obscuring how SSL training governs the structure of learned embeddings. 
To stay clear of such effects, we deliberately opt for a bare-bones unsupervised clustering-based segmentation approach over complex pipelines tuned for SOTA segmentation performance~\cite{wang2023tokencut, wang2023cut,seong2023leveraging, hamilton2022unsupervised}.
While the latter do attain stronger quantitative results, they introduce extensive empirical design choices and hyperparameter spaces, injecting confounding factors into the analysis. 
Our method of choice is \emph{cluster-based probing} by means of across-image k-means clustering, as detailed below.
Despite its simplicity, it forms the core of several SOTA approaches~\cite{hamilton2022unsupervised,hahn2024boosting}.


\vspace{3pt}
\noindent\textbf{Clustering-based probing: }
To directly probe the structure of pretrained embeddings, we use batch-wise clustering as in~\cite{hamilton2022unsupervised,hahn2024boosting}. 
Given a batch of images $\{\mathbf{x}_i\}_{i=1}^{B}$, we extract patch-level embeddings from a frozen SSL backbone, yielding feature embeddings $\mathbf{F}_i^{(l)} \in \mathbb{R}^{N \times d}$, where $N = H_p \times W_p$ is the number of patches, $d$ is the embedding dimension, and $l$ denotes the selected layer. 
We fit a codebook of $K$ cluster centroids $\mathcal{C} = \{\mathbf{c}_k\}_{k=1}^{K} \subset \mathbb{R}^{d}$ to all patch embeddings across the batch via online k-means. 
Each embedding is assigned to its nearest centroid under cosine similarity, and the centroids are updated iteratively across batches, allowing us to capture cross-image semantic structure.
%
See Appendix~\ref{suppl:subsec:unsup_sem_seg_training} for details on hyperparameters and PCA-based initialization of centroids.

%

To produce pixel-wise semantic segmentation predictions, the patch-level embeddings 
$\mathbf{F}_i^{(l)} \in \mathbb{R}^{N \times d}$ are bilinearly upsampled to the 
original image resolution, yielding dense feature maps 
$\hat{\mathbf{F}}_i \in \mathbb{R}^{H \times W \times d}$. 
Each pixel is then assigned a semantic label via nearest-centroid lookup under cosine similarity:
\begin{equation}
    \hat{y}_{hw} = \arg\max_{k \in \{1,\dots,K\}} 
    \frac{\hat{\mathbf{f}}_i^{(hw)} \cdot \mathbf{c}_k}
         {\left\|\hat{\mathbf{f}}_i^{(hw)}\right\|_2 \left\|\mathbf{c}_k\right\|_2},
    \label{eq:assignment}
\end{equation}
where $\hat{\mathbf{f}}_i^{(hw)} \in \mathbb{R}^{d}$ denotes the feature vector at 
spatial location $(h, w)$. 

Taken together, this approach provides a simple, interpretable, and semantically meaningful segmentation of images without introducing any learned decoder, complex post-processing, or iterative self-training.

\subsection{Visualization of Results}
\label{subsec:visualization}
%
\noindent\textbf{Sets of Exemplary Images: }
We operate a standard unsupervised semantic segmentation setting, running 
Hungarian matching globally over the whole test set to map predicted clusters 
to ground-truth labels, and colorizing them respectively using 
the standard dataset color palette. See \cref{fig:protocol}(a1) for examples.

\vspace{3pt}
\noindent\textbf{Aggregate View: }
We construct 
an aggregated map from the class probability map 
$\mathbf{W} \in \mathbb{R}^{K \times H \times W}$, where $\mathbf{W}_{k,i,j}$ 
represents the average predicted probability of class $k$ at pixel $(i, j)$ 
over the entire test set. The per-pixel average probabilities are first normalized as 
$\tilde{\mathbf{W}}_{k,i,j} = \mathbf{W}_{k,i,j} / \sum_{k'} \mathbf{W}_{k',i,j}$, 
and the aggregated map $\mathbf{A} \in \mathbb{R}^{H \times W \times 3}$ is 
then computed as a weighted blend of the standard class colors $\mathbf{v}_k$ of this dataset as 
$\mathbf{A}_{i,j} = \sum_{k=1}^{K} \tilde{\mathbf{W}}_{k,i,j} \cdot \mathbf{v}_{k}$.
%
See \cref{fig:protocol}(a2) for examples.

\subsection{Quantitative Measures}
\label{subsec:quantitative_measures}
\subsubsection{Positional Effect: }
To quantify positional effect in segmentation maps, we compute the class-wise mutual information (MI) $I_c(P;C)$ as follows:
\begin{equation}
    I_c(P;c)=\sum_{p}\mathbb{P}(p,c)\log\frac{\mathbb{P}(p,c)}{\mathbb{P}(p)\mathbb{P}(c)}
\end{equation}
where $P$ denotes the discrete spatial position random variable with realization $p$, and $c$ denotes a specific predicted class instance, and the empirical joint distribution $\mathbb{P}(p,c)$ is estimated from binarized pixel counts over all samples. The global MI and its normalized variant, normalized mutual information (NMI), are then defined as $I(P,C)=\sum_{c} I_c(P;c)$ and $I^\prime(P,C)=\frac{I(P;C)}{H(C)}$, respectively, where $H(C)=-\sum_{c} \mathbb{P}(c)\log(\mathbb{P}(c))$ is the entropy of the predicted class distribution. 
The global NMI summarizes positional bias across all predicted classes, and enables comparability across embedding types and model families. 
See \cref{fig:protocol}(b1) for examples.
Optionally, the class-wise MI, visualized for highest-scoring classes, may convey further insightful context. See \cref{fig:scaling_different_models}(b3) for examples. 

\vspace{3pt}
\noindent\textbf{Segmentation Quality: } Following standard practice for unsupervised semantic segmentation evaluation, 
we employ mean Intersection-over-Union (mIoU) computed under optimal label permutation via Hungarian matching between predicted clusters and ground-truth labels.
See \cref{fig:protocol}(b2) for examples.

\subsection{Scenarios of Interest}
\label{sec:protocol-scenarios}
%
%
Beyond insights into individual embedding spaces, applying our protocol multiple times 
and putting results next to each other, we can gain a range of \emph{comparative} insights, including the following scenarios of interest:
\begin{itemize}
    \item To observe differences across training paradigms: Pick one architecture, embedding type, and layer selection strategy -- vary the training paradigm;
    \item To observe how model behavior evolves across ViT layers: Pick one model and embedding type -- vary the layer;
    \item To observe differences across model sizes: Pick one training paradigm, embedding type, and layer selection strategy -- vary the model size.
\end{itemize}
Regarding layer selection strategies, we can straightforwardly pick identical layers given a fixed architecture. Alternatively, we can pick layers with corresponding properties, like e.g.\ for any given model and embedding type, pick the layer that achieves best unsupervised semantic segmentation performance in terms of mIoU. 
The latter serves the specific purpose of gaining insights into the extent to which, for any given model, its embedding space with best emerging semantic structure is confounded by positional effect. Furthermore, it has the additional advantage that it is applicable for fixed as well as varying model architecture.

%% file: content/4_experiments.tex
\section{Results and Discussion}
\label{sec:results}
\subsection{Experimental setup}
\subsubsection{Datasets: }  
A dataset ideally suited for model understanding by means of our protocol would have no \emph{inherent} spatial bias, as dataset-inherent spatial bias acts as a confounder by entangling true semantic structure with positional effect. Furthermore, a suitable dataset necessarily needs to allow for meaningful unsupervised semantic segmentation benchmarking of SSL ViTs. 
We use COCO-Stuff~\cite{caesar2018coco} as the main dataset 
because it has the least dataset-inherent spatial bias among common unsupervised segmentation benchmarks. 
In particular, it features diverse scene layouts and respective weaker spatial-semantic correlations than a number of other, more object-centric benchmarks. It further features diverse background class labels, allowing us to assess a model's ability to distinguish semantic background content (e.g., sky, grass, ceiling), not just foreground objects (e.g., people, animals).
Following \cite{hahn2024boosting,hamilton2022unsupervised, seong2023leveraging}, we chose a 27-class subset of COCO-stuff.
That said, to ensure that our study does not merely reveal COCO-Stuff-inherent behaviors or hidden biases, we additionally evaluated on Cityscapes~\cite{cordts2016cityscapes} and the PascalPart~\cite{chen2014detect} animals subset, 
see Appendix~\ref{app:further_datasets}.

\vspace{3pt}
\noindent\textbf{SSL models: }
All models in our study share a Vision Transformer (ViT) backbone~\cite{dosovitskiy2020image}, enabling architecture-aligned comparisons.
We evaluate eight representative SSL frameworks -- MAE~\cite{he2022masked}, MoCov3~\cite{chen2021empirical}, Mugs~\cite{zhou2022mugs}, iBOT~\cite{zhou2021ibot}, DINO~\cite{caron2021emerging}, DINOv2~\cite{oquab2023dinov2}, including its register-token variant~\cite{darcet2023vision}, and DINOv3~\cite{simeoni2025dinov3} -- and two supervised baselines: a standard ImageNet-supervised ViT~\cite{dosovitskiy2020image} and  CLIP~\cite{ilharco2021openclip}.
ViT-Base serves as the primary architecture, with ViT-Large included to assess scaling effects.
%
Key characteristics of each model are listed in Appendix~\ref{app:model-properties}.

\vspace{3pt}
\noindent\textbf{Experiments: } 
We run the following setups of our protocol: 
\begin{itemize}
\item We vary the model training paradigm, at fixed ViT-B architecture and fixed layer selection policy based on best segmentation performance. 
\item We vary the layer (from first to last), for some fixed ViT-B models. 
\item We vary model size at fixed training paradigm and fixed layer selection policy based on best segmentation performance. 
\end{itemize}
%



%
\subsection{Results}
\label{subsec:results}
\begin{figure}[t]%
    \centering
    \includegraphics[width=\textwidth]{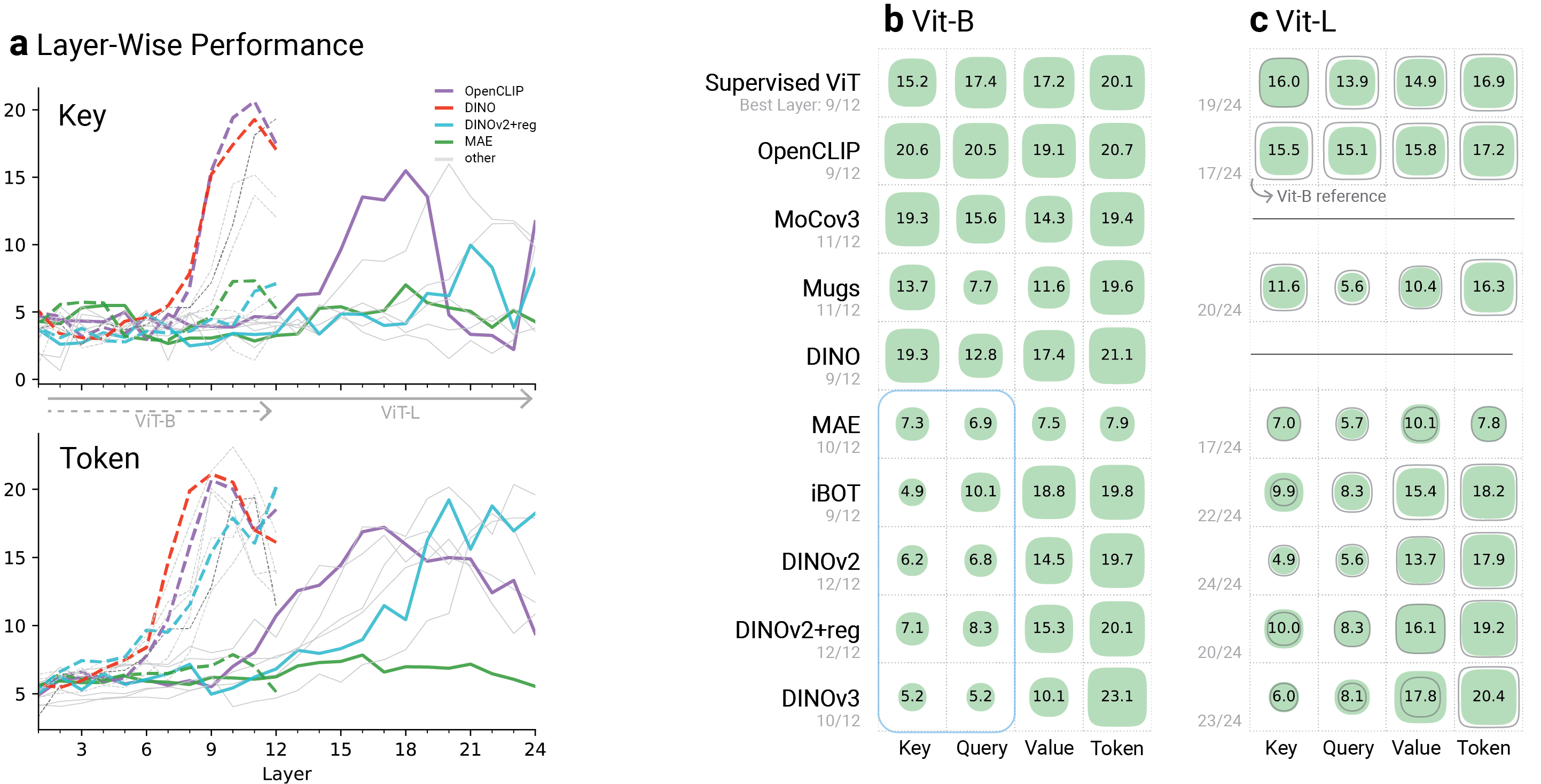} %
    \caption{Performance of \textbf{key}, \textbf{query}, \textbf{value}, and \textbf{token} embeddings in the \textbf{unsupervised semantic segmentation} task across SSL models and model sizes. \textbf{a)} displays the layer-wise segmentation performance based on key and token embeddings for four selected models. \textbf{b)}/\textbf{c)} present the overall performance of all models, using the respective best-performing layer for ViT-Base and ViT-Large. Models highlighted in blue exhibit notably low segmentation performance with query and key embeddings.}%
    \label{fig:fig_performance_task_2}%
\end{figure}

Unless otherwise specified, results presented here are for ViT-B models on COCO-Stuff. 
Corresponding ViT-L results as well as additional models and datasets omitted here for clarity are provided in Appendices~\ref{app:unsup_seg_qual_result},~\ref{app:unsup_over_seg} and~\ref{app:further_datasets}. 
Regarding layer selection by best segmentation performance, \Cref{fig:fig_performance_task_2} lists the best-performing layer for each model and embedding type. 
Note that we pick last-layer tokens \emph{after} the final LayerNorm; see Appendix~\ref{app:layernorm_effect} for a respective ablation.

\vspace{3pt}
\noindent\textbf{Behavior across Varying Training Paradigms: }
\Cref{fig:protocol} shows results for the key embeddings from the best-performing (in terms of segmentation mIoU) layer of each of several MIM-based models, alongside DINO as a CL baseline. For MIM, as opposed to CL, embeddings cluster consistently across images according to their global location, indicating strongly dominant positional effect. This observed behavior is related yet distinct from previous findings regarding locality bias in MIM vs.\ CL models~\cite{simeoni2025dinov3,amir2021deep,bolya2025perception}. 
We further discuss the relation between positional effect and locality bias in \cref{subsec:insights}(1).

At the same time, MIM keys from peak-performing layers exhibit considerably inferior segmentation performance (mIoU) as compared to CL, see also \cref{fig:fig_performance_task_2}: Here, MIM keys and queries form a subset (highlighted by blue outline) of notably lower performance compared to other models and embedding types. 
This gives rise to the hypothesis that pronounced positional effect causes degraded segmentation performance (see \cref{subsec:insights}(2) for further discussion). 
%

Deserving of mention is the fact that observable positional effect is already inherent in the ground-truth (GT) semantic labels themselves, arising from characteristic spatial regularities in natural image statistics. E.g., \textit{humans} are disproportionately concentrated near the image center, \textit{sky} occupies the upper region, and \textit{ground} predominantly appears in the lower half. Nevertheless, the positional effects in MIM keys substantially exacerbate this tendency, as revealed by respective aggregate visualizations and quantitative measures (cf.\ \cref{fig:protocol}(a2) and (b1)): Across all models we observe elevated global NMI relative to GT. 

As a side note, block-structured artifacts are clearly visible in DINOv2 results, which have been reported and attributed to positional embeddings in~\cite{yang2024denoising}. See Appendix~\cref{app:unsup_seg_qual_result} for further respective results, also including CLIP.

\vspace{3pt}
\noindent\textbf{Behavior across Layers: }
\begin{figure*}[p]
    \centering
    \begin{subfigure}{1\textwidth}
    \centering
    \includegraphics[width=0.9\columnwidth]{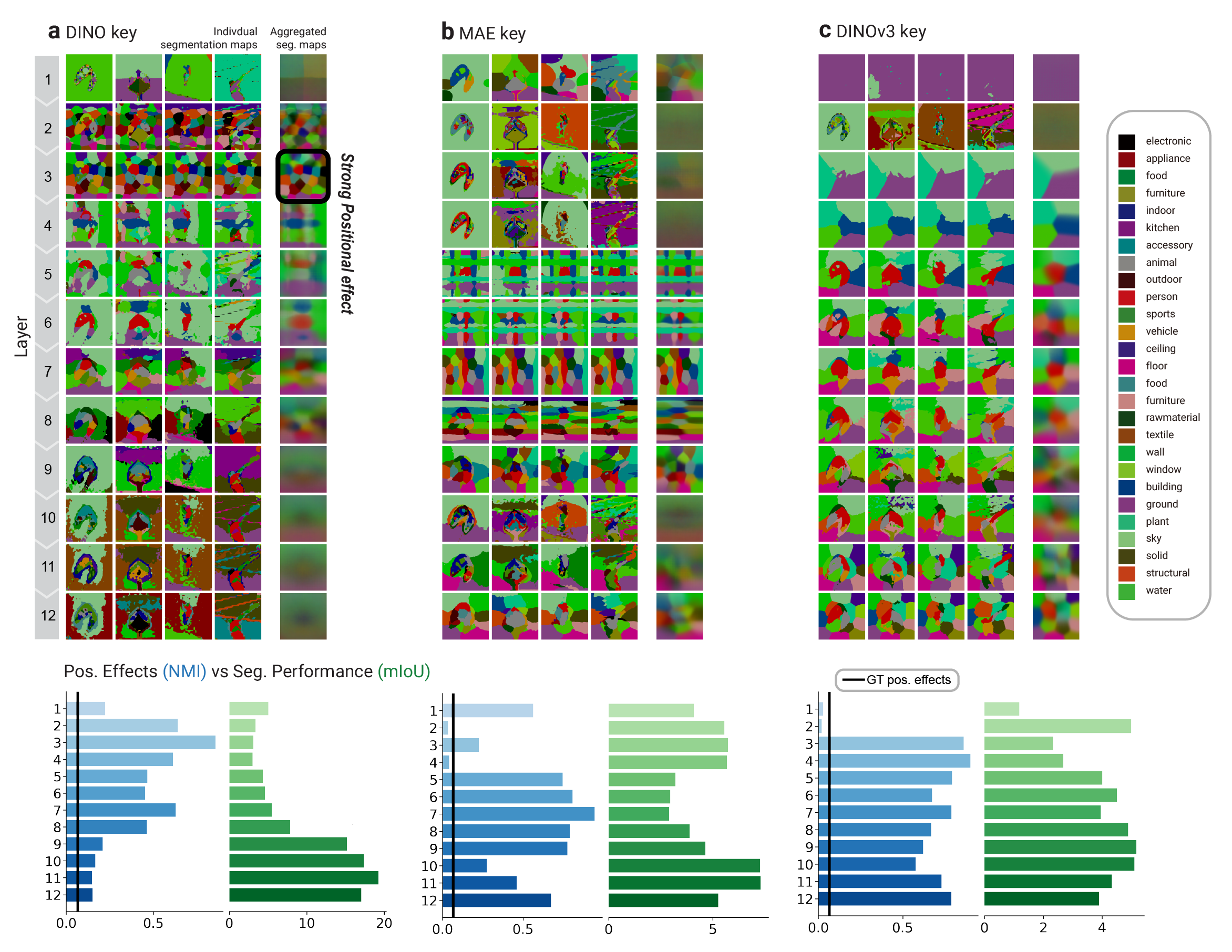}
    \end{subfigure}
\hrule

    \begin{subfigure}{1\textwidth}
    \centering
    \includegraphics[width=0.9\columnwidth]{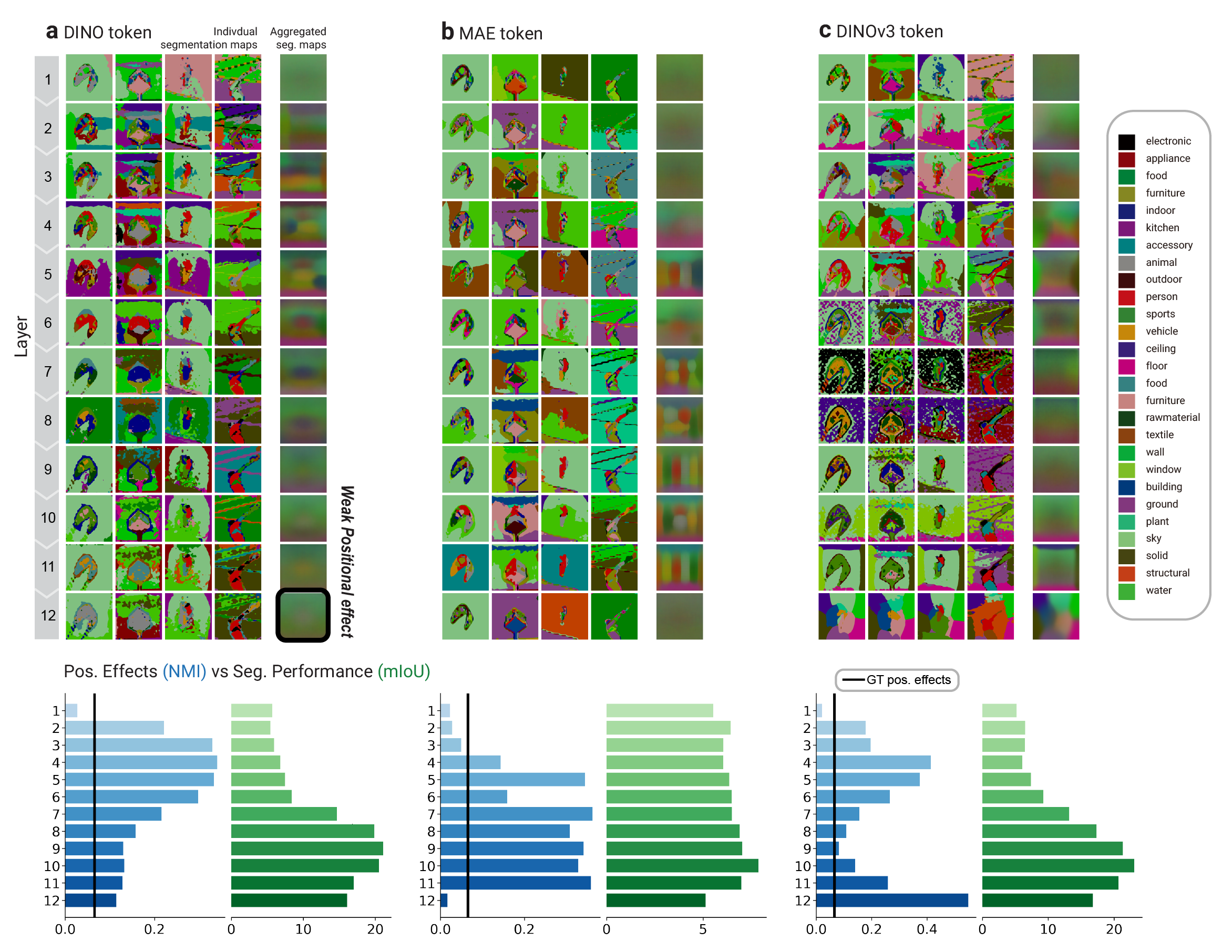}
    \end{subfigure}
    \caption{\textbf{Model behavior across layers:} Unsupervised semantic segmentation results for three exemplary models: \textbf{a)} DINO, \textbf{b)} MAE, and \textbf{c)} DINOv3. 
    \textbf{Top: Key embeddings:} The three models exhibit distinct patterns of positional effect: it is confined to earlier layers in DINO,  strongly emerges in intermediate layers in MAE, and persists strongly across all layers in DINOv3.
    \textbf{Bottom: Token embeddings: } 
    Positional effect is largely reduced compared to key embeddings across  models. Notably, however, DINOv3 remains an exception, with its last layer still exhibiting strong positional effect. 
    }
    \label{fig:pos_bias_layer_wise}
\end{figure*}
\Cref{fig:pos_bias_layer_wise} provides results across layers for keys (top) and tokens (bottom) of DINO, MAE and DINOv3. 
Regarding keys, DINO exhibits positional effects in early layers (layers 2 and 3), which diminish progressively in deeper layers. In contrast, MAE shows strong positional effects only in later layers (layer 5), while DINOv3 is dominated by positional effect consistently across all layers. 
%
%
Regarding tokens, positional effect is substantially reduced compared to keys across all models; DINOv3 remains a notable exception, as its last layers exhibit a sharp onset of strong positional effect.

DINOv3's comparatively strong positional effects across embeddings and layers is surprising as its use of RoPE makes positional embeddings relative as opposed to absolute -- thus we would expect them to manifest as locality bias rather than positional effect, or at least expect diminished positional effects compared to models not using RoPE -- yet we observe the opposite.

\begin{figure}[htbp]%
    \centering
    \includegraphics[width=0.97\textwidth]{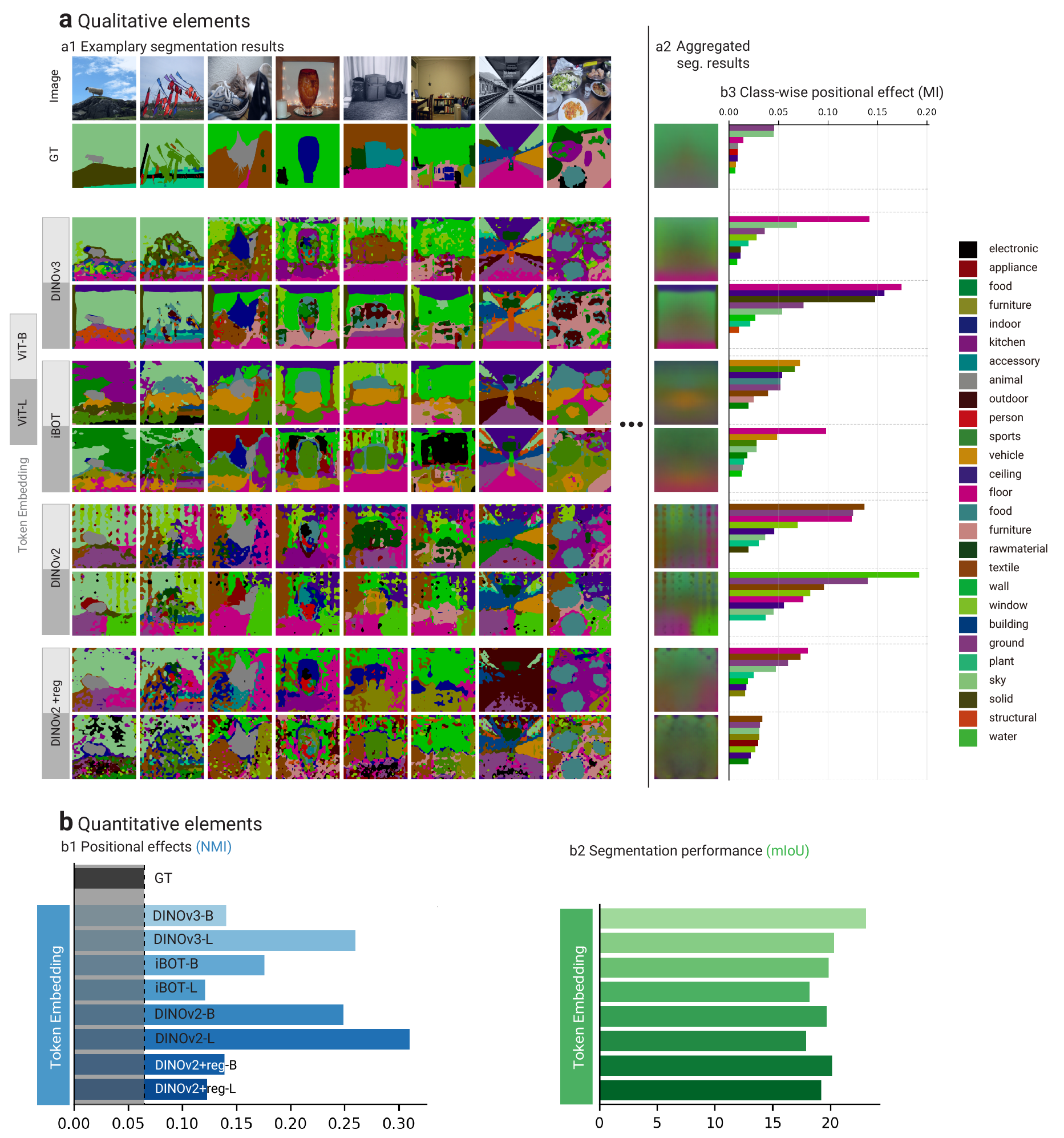} %
    \caption{\textbf{Behavior across scales:} Clustering results for \textbf{token embeddings} from the best-performing layer of the Base and Large variants of DINOv3, iBOT, DINOv2 and DINOv2+reg.
    We observe a notably increased positional effect in DINOv3-L compared to its -B counterpart alongside inferior segmentation performance in terms of mIoU. DINOv2 exhibits a similar pattern. 
    For iBOT, we observe notable positional effects across sizes, yet to a lesser extent and along with a shift in clusters in the -L variant, as further studied in \cref{fig:scaling_ibot_body_parts}.
    For DINOv2 with registers, increasing model size leads to a larger number of small, fragmented clusters, consistent with observations in \cite{simeoni2025dinov3}.}%
    \label{fig:scaling_different_models}%
\end{figure}

\vspace{3pt}
\noindent\textbf{Scaling Behavior: } 
\Cref{fig:scaling_different_models} provides results across scales, namely ViT-B vs.\ ViT-L, for DINOv3, iBOT, DINOv2 and DINOv2+reg token embeddings. 
We highlight a particularly notable finding concerning the positional effect in DINOv3-Large: Here, individual and aggregated segmentation maps reveal that the upper and lower regions of images are heavily affected by positional effect, forming spurious clusters systematically mis-assigned as \textit{ceiling} and \textit{floor} irrespective of ground-truth class. This positional effect is also partially visible along vertical boundaries, where affected regions are incorrectly segmented as a single \textit{solid} class. In comparison, the DINOv3-Base variant, while also affected, exhibits less pronounced positional effect, only in the lower boundary region. Accordingly, DINOv3-Large shows disproportionately high class-wise MI values (cf.\ \cref{fig:scaling_different_models}(b3)) for semantically spurious classes such as \textit{floor}, \textit{ceiling}, \textit{solid}, as well as high global NMI as compared to its Base counterpart. We hypothesize that this pronounced positional effect in the Large variant is a key contributor to DINOv3's degraded scaling behavior, as discussed in more depth in \cref{subsec:insights}.

\begin{figure}[htbp]%
    \centering
    \includegraphics[width=0.97\textwidth]{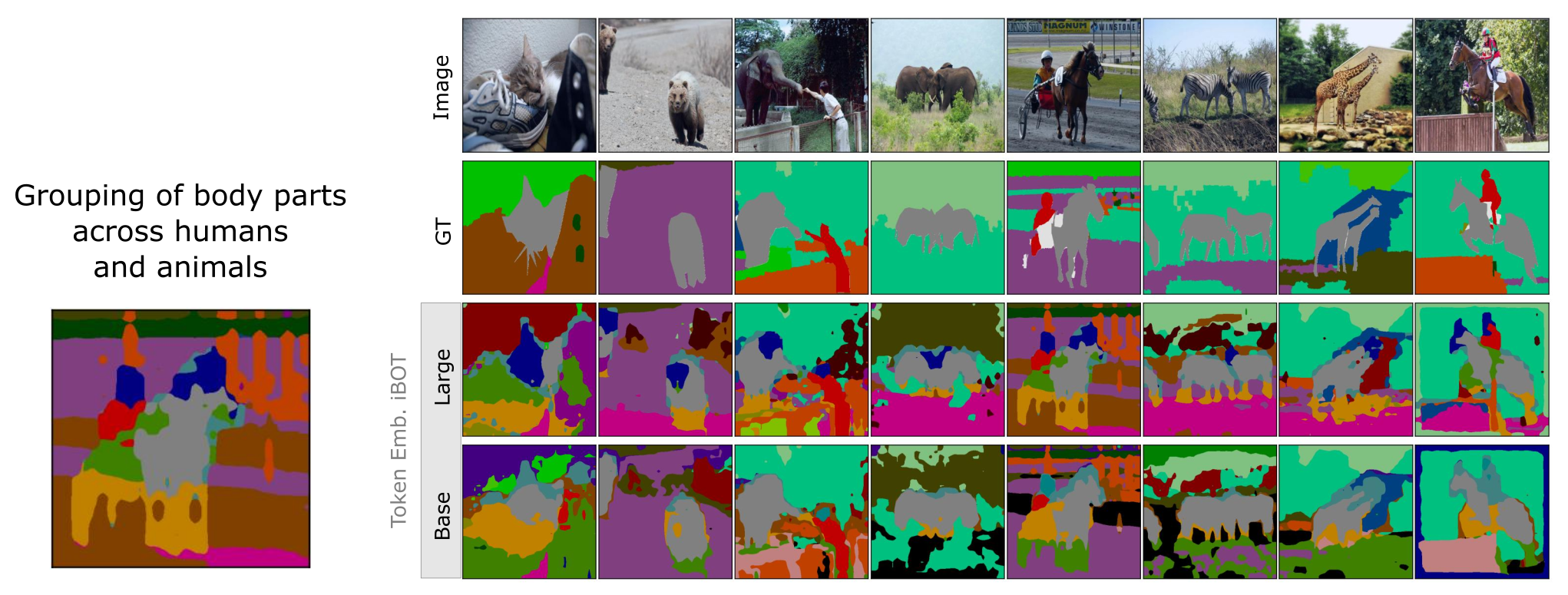} %
    \caption{Positive scaling effect in iBOT: Unsupervised segmentation results for token embeddings from the best-performing layer of the Base and Large variants of iBOT focusing on animal images. We observe more fine-grained clustering of object parts for the larger model. For example, in the Large model, head regions of humans and different animals are assigned to more similar embeddings compared to the Base variant.}%
    \label{fig:scaling_ibot_body_parts}%
\end{figure}

Beyond DINOv3, we observe further model-specific patterns that become more pronounced with scale, see~\cref{fig:scaling_different_models}:
%
%
DINOv2-Large displays a similarly strong positional effect as DINOv3, manifesting as consistent cluster assignments tied to fixed spatial regions across images. 
In contrast, DINOv2+reg largely mitigates this  effect. However, it produces noticeably noisier segmentations, though boundary localization itself does not appear to degrade. This increased noise aligns with  patch inconsistencies reported in~\cite{simeoni2025dinov3} for larger models.

Regarding iBOT, both Base and Large variants exhibit pronounced positional effects; see e.g.\ the recurring appearance of a "vehicle-like" segment in the lower central region of images irrespective of semantic content. However, cluster boundaries still tend to respect the underlying semantic structure of the images.
Besides positional effect, we also notice that iBOT exhibits more fine-grained and semantically consistent clustering as model size increases: In the Large variant, body parts such as heads, legs, and arms are more consistently grouped across humans and animal species compared to the Base model (see also~\cref{fig:scaling_ibot_body_parts}).
Increasing the number of clusters further reveals more fine-grained, position-aware partitions of these body parts as well as other objects (see Appendix~\ref{app:unsup_over_seg}).

In summary, as observed quantitatively in~\cref{fig:fig_performance_task_2}, increasing model size does not necessarily translate into improved performance in downstream semantic segmentation. 
In many cases, Large models underperform their Base counterparts, consistent with observations in~\cite{bolya2025perception,simeoni2025dinov3}. 
Note that this is in contrast to zero-shot semantic keypoint correspondence, where we observe favorable scaling behavior (cf.\ Appendix~\cref{app:semantic_correspondence}), similar to reports for depth estimation~\cite{dehghani2023scaling}. 
While~\cite{balestriero2025lejepa,simeoni2025dinov3} attribute inverse scaling to a pretext/downstream task misalignment and loss of patch consistency, respectively, we observe increased positional effect as an additional, previously unreported contributor, in particular for DINOv3.



\subsection{Extended Insights and Discussion}
\label{subsec:insights}
\subsubsection{(1) Positional effect causes known locality bias in MIM: }
%
\begin{figure}[p]%
    \centering
   
    \includegraphics[width=0.95\textwidth]{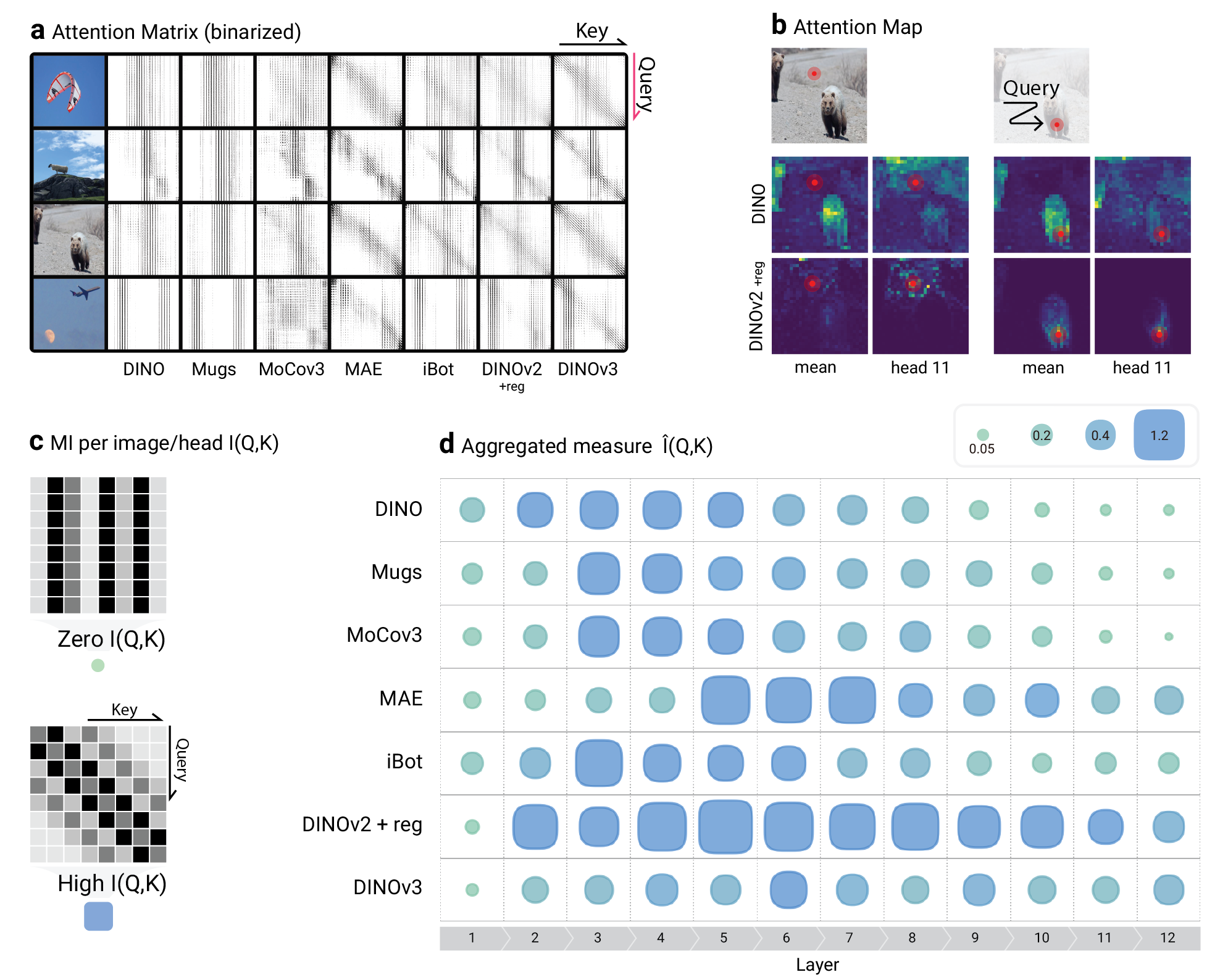} %
    \caption{\textbf{Locality Biases} observable in Attention Matrices. \textbf{a)} Attention matrices for multiple models, averaged over attention heads. For improved visualization, maps were binarized w.r.t.\ their 95th percentile. \textbf{b)} Exemplary attention maps for two different queries, illustrating the stronger locality of attention in DINOv2+reg compared to DINO. \textbf{c)} Vertically “barcoded” attention patterns indicate zero mutual information $I(Q,K)$ between respective keys and queries, whereas diagonal structures correspond to high $I(Q,K)$. \textbf{d)} Aggregated measure $\hat{I}(Q,K)$, averaged over images and heads. Notably, CL models (DINO, Mugs, MoCov3) exhibit a pronounced decrease in $\hat{I}(Q,K)$ towards deeper layers, while MIM models maintain a higher mean. }%
    \label{fig:barcodes}%
\end{figure}
%
Previous works have repeatedly shown the existence or diminishing of \emph{locality bias} across layers~\cite{simeoni2025dinov3,amir2021deep,bolya2025perception}, also revealing respective distinct behaviors of MIM vs.\ CL. We here replicate and expand upon these findings (see Fig.\ \ref{fig:barcodes}), identifying positional effects in keys and queries as an upstream cause for known locality biases. 

In more detail, it has been shown that the vertical stripes in attention matrices (see Fig.\ \ref{fig:barcodes}(a)) reflect two phenomena:
If the stripe corresponds to a background position, this indicates the use of background locations to store global information~\cite{darcet2023vision}.
If the stripe corresponds to a foreground position, this has been termed \emph{query collapse} and indicates that even most background queries attend to the foreground object~\cite{park2023whatssllearn}.
The diagonals, on the other hand, reflect locality bias -- i.e., independent of position, queries attend to nearby positions~\cite{bolya2025perception}.

To compare these effects quantitatively across models/layers, we borrow the information-theoretic notion of MI~\cite{shannon1948mathematical}, applied to query--key attention maps; i.e., we compute the MI $I(Q,K)$ per image/head and aggregate to yield $\hat{I}(Q,K)$, both defined in Appendix~\ref{app:locality_bias_mi} \crefrange{eq:locality-bias-proxy-per-image}{eq:locality-bias-proxy}. This proxy measure of locality bias yields higher values for diagonal/local query--key structures (see Fig. \ref{fig:barcodes}(c)).

%
We can thus quantitatively confirm that locality bias decreases for CL models in deeper layers.
For MAE it emerges at mid-depth and stays elevated until the end (see~\cref{fig:barcodes}(d)).
DINOv2+reg shows the strongest and most persistent locality bias; DINOv3 has a similarly persistent bias but at a lower level.
\begin{figure}[p]%
    \centering
    \includegraphics[width=0.95\textwidth]{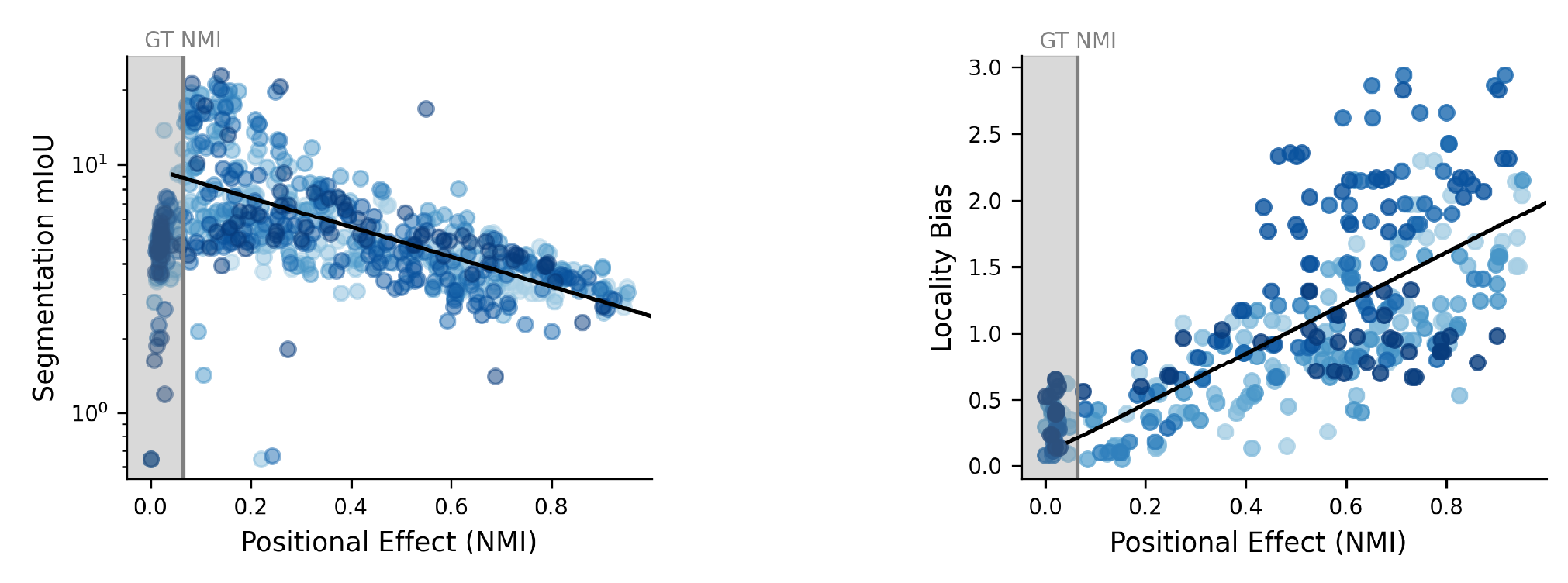} %
    \caption{\textbf{Left:} Correlation between positional effect and segmentation performance (in log-scale) for all models and layers, across Keys, Queries, Values, and Tokens. \textbf{Right:} Correlation between positional effect and locality bias in Key and Query Embeddings.}%
    \label{fig:correlation}%
\end{figure}

As expected, we find strong correlation between key/query positional effect and locality bias in terms of $\hat{I}(Q,K)$ ($\rho =0.70$ with $p<.001$, see \cref{fig:correlation} (\textbf{right})). We deem consistent positional effects in keys and queries at any given layer causal for downstream locality bias: Attention values that serve to quantify locality bias are directly computed as (normalized) dot products of their upstream keys and queries; thus downstream locality bias directly follows.

\vspace{3pt}
\noindent\textbf{(2) Positional effect correlates negatively with segmentation performance:}
 We quantified the correlation between positional effect and segmentation mIoU, finding significant negative correlation ($\rho= -0.59$ with $p<.001$, see \cref{fig:correlation} (\textbf{left})). Regarding respective causality: Any delta positional effect over the ground truth labels' effect constitutes segmentation error; however, not all segmentation error is due to (adverse) positional effect. We hypothesize that the observed inverse scaling behavior of DINOv3 can be attributed to the measured increased positional effect because our visualization suggests notable impact; yet there are confounders we cannot disentangle, e.g., potentially shifting (mis-) alignment of model-encoded clusters and GT labels (as discussed in \cref{subsec:results}).

\vspace{3pt}
\noindent\textbf{(3) Tokens yield highest segmentation performance: }
Our quantitative evaluation of segmentation performance in~\cref{fig:fig_performance_task_2} shows that token embeddings consistently achieve the highest performance across models (the only exception being MAE-Large). The same trend is observed for the semantic correspondence task (see Appendix~\ref{app:task1_layerwise}).
This finding contrasts with the results of~\cite{amir2021deep}, who report superior performance of key embeddings in keypoint matching. We attribute this discrepancy primarily to the limited model scope of \cite{amir2021deep}, which evaluates exclusively on DINO, a model where the performance gap between key and token embeddings is relatively small, and on only 360 sampled image pairs. 
%
%
Similarly, prior works on unsupervised segmentation pipelines (e.g., TokenCut and CutLER~\cite{wang2023cut,wang2023tokencut}) report that key embeddings from DINO can outperform token embeddings. However, these methods address per-image object discovery rather than dataset-level semantic segmentation: Their graph-based clustering is performed \emph{independently for each image}. Hence their conclusions apply to object- or region boundaries, yet not necessarily to \emph{across-image} semantic classes. 

At the same time, our layer-wise analysis of unsupervised semantic segmentation in~\cref{fig:fig_performance_task_2} and Appendix  \cref{fig:task2_kqvt,fig:task1_kqvt} (for the keypoint correspondence task) confirms previous findings that, across model families, optimal downstream performance most often emerges in intermediate rather than the final layers~\cite{bolya2025perception,simeoni2025dinov3}.

%% file: content/5_conclusion.tex
\section{Conclusion}

This work proposes a simple and straightforward visualization protocol for intuitive understanding of ViT representations. We employ it in a systematic benchmark of modern self-supervised vision models, across layers and embedding types. Evaluating eight representative SSL models together with supervised and CLIP baselines, we provide a unified and architecture-controlled analysis of how different pretraining objectives shape ViT representations.

Our benchmark yields several previously unreported insights on top of intuitively and consistently visualizing a broad range of previously reported findings. Regarding novel insights, we find that \textbf{dominant positional effect in keys and queries causes locality bias} previously observed for Masked Image Modeling. Furthermore, \textbf{inverse scaling of DINOv3 appears to stem from strong positional effect}, calling for a thorough reassessment of its training paradigm (including distillation) as well as its use of RoPE. 


We release all code and results, 
serving as a resource for understanding \emph{where} and \emph{how} semantic information and positional effect emerge in large-scale vision pretraining, thus facilitating further model understanding and advancement.





\vspace{6pt}\noindent\textbf{Acknowledgments: }
This work was supported by the Helmholtz Einstein International
Berlin Research School in Data Science (HEIBRiDS),  German Research
Foundation (DFG) Research Training Group CompCancer
(RTG2424), DFG Collaborative Research Center FONDA (SFB 1404, project no. 414984028), and the Synergy Unit of the Helmholtz Foundation Model Initiative.

%% file: content/7_appendix.tex
\newpage
\section{Appendix}
\subsection{Properties of models included in the benchmark}
\label{app:model-properties}
\input{tabs/camera_ready_version/ssl_table_1}
\input{tabs/camera_ready_version/ssl_table_2}

\subsection{Unsupervised semantic segmentation training setup and sensitivity analysis}
\label{suppl:subsec:unsup_sem_seg_training}
We minimize the batch-wise k-means loss~\cite{mcqueen1967some} using cosine similarity as the distance metric, training for a fixed 5{,}000 steps with the Adam optimizer (learning rate $5 \times 10^{-3}
$, batch size 24).

To initialize the cluster centroids, we adopt a PCA-based strategy tailored to our unsupervised semantic segmentation experiments on COCO-Stuff-27. Specifically, PCA is performed on patch embeddings sampled from the training set, and the top $K = 27$
principal components are used as the initial centroids. By default, patches are drawn from the first 96 batches, corresponding to approximately 2,300 images. 

\begin{figure}[htbp]%
    \centering
    \includegraphics[width=\columnwidth]{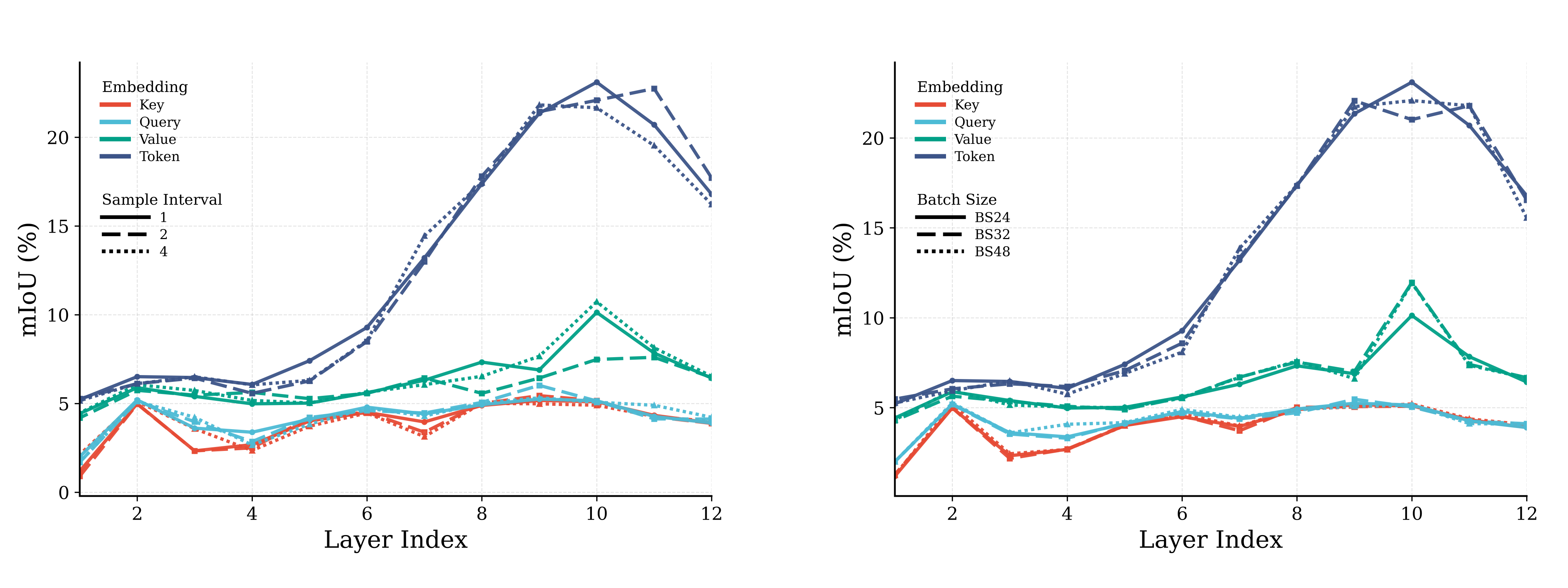} 
    \caption{\textbf{Left:} Impact of PCA-based centroid initialization on final inference performance for DINOv3 Base model. Results are reported across three different subsets of training patches used for PCA initialization. \textbf{Right:} Impact of batch size on final inference performance for DINOv3 Base model.}
    \label{fig:pca_init}%
\end{figure}

To assess the robustness of the initialization, we jointly vary the number of sampled batches and the sampling interval while keeping the total number of patches used for PCA initialization constant. Specifically, we evaluate configurations with 2× and 4× more batches, paired with sampling intervals of 2 and 4, respectively.
Furthermore, we additionally experiment with batch sizes of 32 and 48, compared to the default batch size of 24 to evaluate sensitivity to batch size. We report the final training performance only for the DINOv3 Base model across these configurations in order to quantify the variance of final segmentation performance attributable to these hyperparameters in \cref{fig:pca_init}.

\subsection{Locality bias measured by mutual information
\label{app:locality_bias_mi}}

To quantify locality bias, we borrow the information-theoretic notion of conditional mutual information and apply it on the query-key attention maps. For each image and attention head, we interpret the normalized patch-to-patch attention weights as a conditional distribution over key positions given a query position, and compute the resulting mutual information between query and key positions. Averaging these values over images and heads yields an empirical estimate of $\widehat I_{\mathcal D}(Q;K\mid X,H)$, which we use as an attention-based proxy for locality bias.

Let $\mathcal{D}=\{x_m\}_{m=1}^{M}$ denote the dataset, and let $A^{(m,h)}\in\mathbb{R}^{N\times N}$  be the patch-to-patch attention matrix for an input image $x_m$ and head $h$, after excluding the CLS token. We use the convention that rows correspond to query positions and columns to key positions, so that $A^{(m,h)}(q,k)$ denotes the attention weight from query patch $q$ to key patch $k$.

Upon removal of the CLS token, the attention matrix after the softmax is already row normalized. Preserving this, we renormalize each query row over patch keys: $\tilde{A}(q,k)
= A(q,k) / \sum_{\ell=1}^{N}A(q,\ell)
$ (dropping $m$, $h$ for simplicity) Hence, $\tilde{A}^{(m,h)}(q,k)$ defines an attention-induced conditional distribution
\[
\mathbb{P}(K=k\mid Q=q,X=x_m,H=h)=\tilde{A}^{(m,h)}(q,k).
\]
Assuming a uniform distribution over query patches, $\mathbb{P}(Q=q\mid X,H)=1/N$, the induced joint distribution is
\[
\mathbb{P}(Q=q,K=k\mid X=x_m,H=h)
=
\frac{1}{N}\tilde{A}^{(m,h)}(q,k).
\]
The marginal distribution over key positions is then given by
\[
\mathbb{P}(K=k\mid  X=x_m,H=h) =  \bar{A}^{(m,h)}(k):=
\frac{1}{N}\sum_{q=1}^{N}\tilde{A}^{(m,h)}(q,k).
\]
We then compute the mutual information~\cite{shannon1948mathematical} between query and key positions given
an image $x_m$ and head $h$:
\begin{equation}
\label{eq:locality-bias-proxy-per-image}
I(Q;K | X = x_m, H = h)
=
\sum_{q=1}^{N}\sum_{k=1}^{N}
\frac{1}{N}\tilde{A}^{(m,h)}(q,k)
\log \left(
\frac{
\tilde{A}^{(m,h)}(q,k)
}{
\bar{A}^{(m,h)}(k)
}\right).
\end{equation}
Note that intuitively $\bar{A}^{(m,h)}(k)$ is the query-agnostic baseline, i.e.,\ how much each key is attended to on average across all queries. So if all query patches attend to the same key distribution, then $\bar{A}^{(m,h)}(k) = \tilde{A}^{(m,h)}(q,k)$ for all $k$ and thus $I_{m,h}(Q;K)
= 0$ (see Fig. \ref{fig:barcodes}(c)). 

Finally, assuming uniform weighting over images and heads, we average the per-image, per-head values to obtain the empirical conditional mutual information 
\begin{equation}
\label{eq:locality-bias-proxy}
 \widehat I_{\mathcal D}(Q;K\mid X,H)
=
\frac{1}{M N_h}
\sum_{m=1}^{M}
\sum_{h=1}^{N_h}
I_{m,h}(Q;K).
\end{equation}

For notational simplicity, we denote the empirical conditional mutual information in the main paper by $\hat{I}(Q,K) := \widehat I_{\mathcal D}(Q;K\mid X,H)$ and equivalently $I(Q,K) := I(Q;K\mid X = x_m,H = h)$.

\subsection{Unsupervised semantic segmentation layer-wise performance}
\label{app:task2_layerwise}
\begin{figure}[htbp]%
    \centering
    \subfloat[\centering key embedding]{{\includegraphics[width=6.4cm]{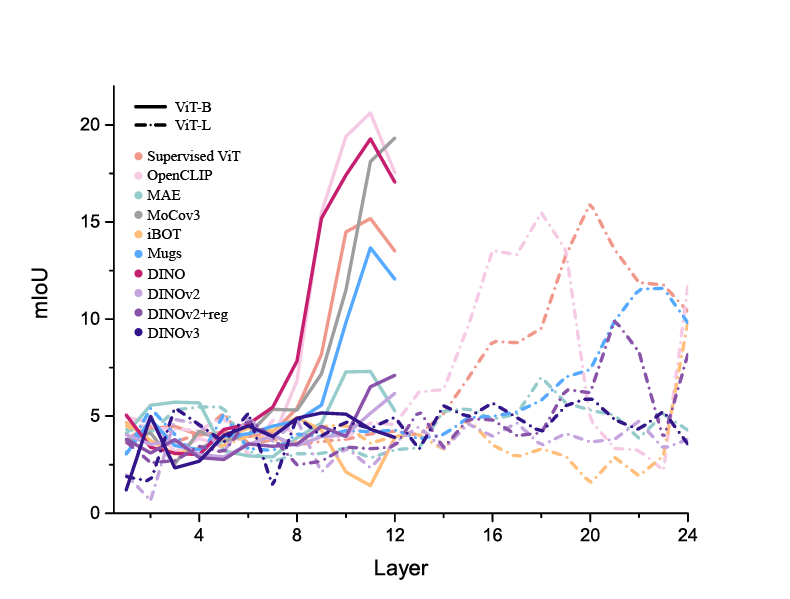} }}%
    \subfloat[\centering query embedding]{{\includegraphics[width=6.4cm]{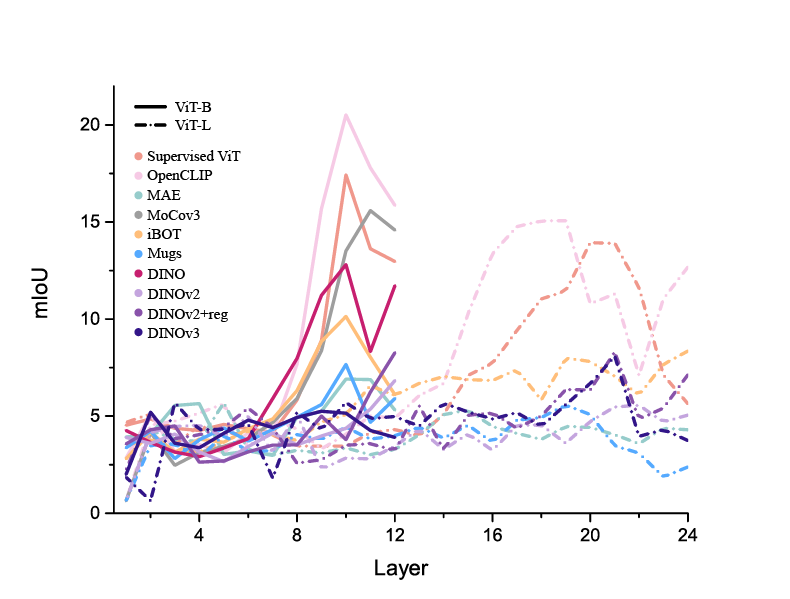} }}%
    \\[0.5em]
    \subfloat[\centering value embedding]{{\includegraphics[width=6.4cm]{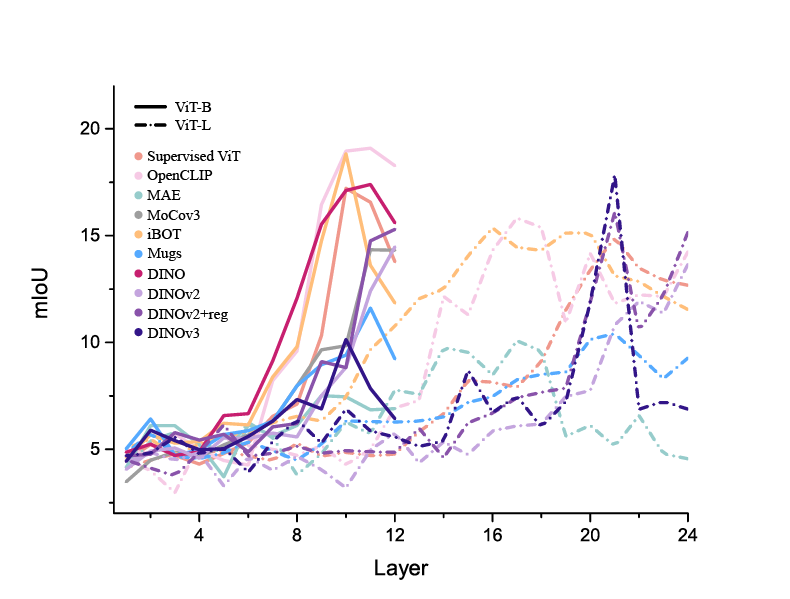} }}%
    \subfloat[\centering token embedding]{{\includegraphics[width=6.4cm]{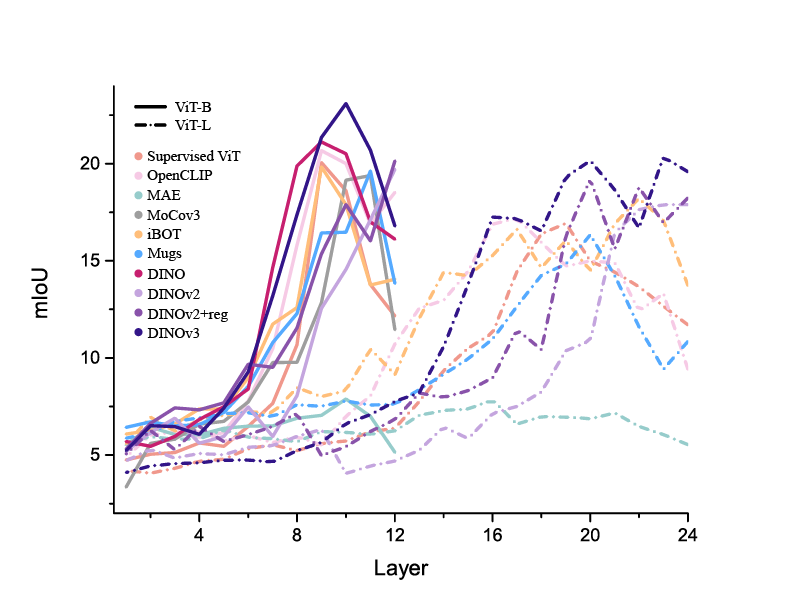} }}%
    \caption{Performance of \textbf{key}, \textbf{query}, \textbf{value} and \textbf{token} embeddings on \textbf{unsupervised semantic segmentation} task across layers and model sizes.}%
    \label{fig:task2_kqvt}%
\end{figure}

\Cref{fig:task2_kqvt} presents the layer-wise performance on unsupervised semantic segmentation for both base and large models across all four embedding types.

\subsection{Unsupervised segmentation results for further models}
\label{app:unsup_seg_qual_result}
\begin{figure*}[ht]
    \centering
    \includegraphics[width=\columnwidth]{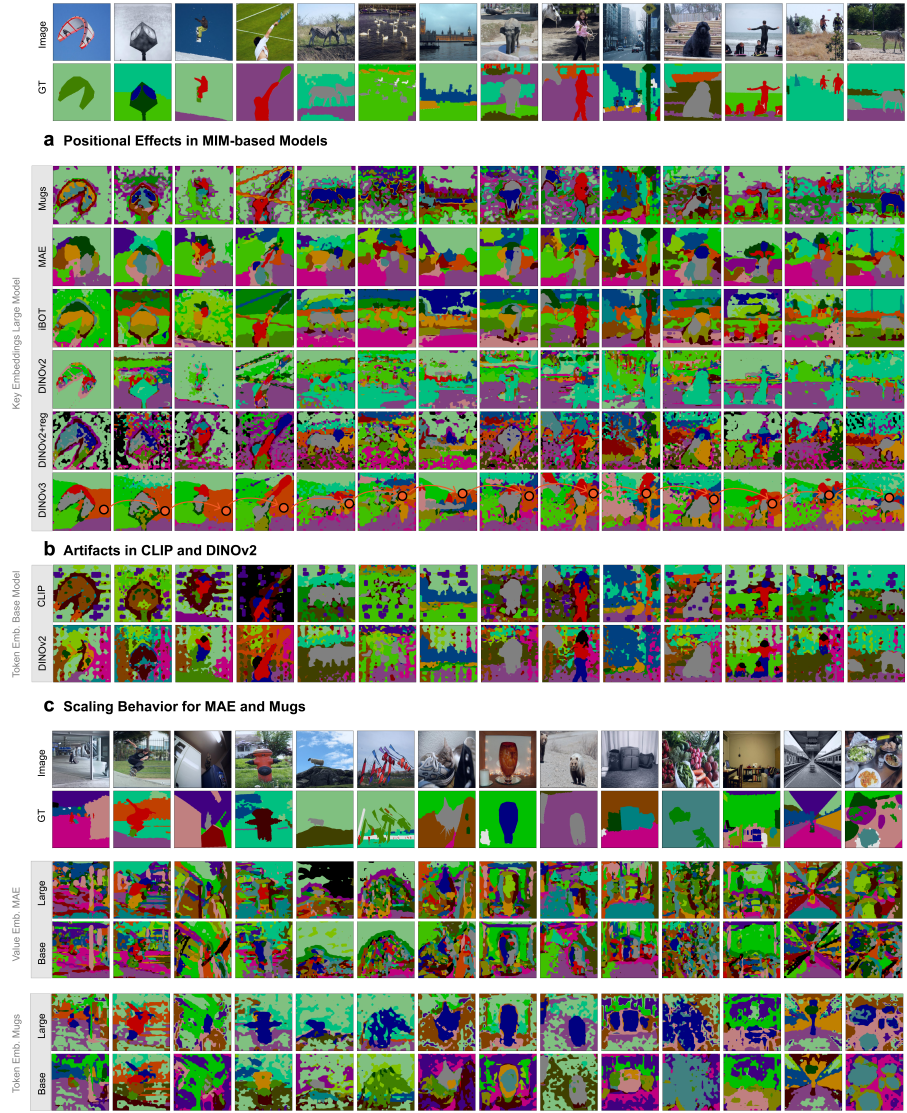}
    \caption{\textbf{Additional examples of unsupervised semantic segmentation across images.} 
    \textbf{a)} Reproduction of ~\cref{fig:protocol} using key features from the Large models.
    \textbf{b)} Positional bias in CLIP and DINOv2, with segmentation results obtained from token embeddings of the Base models.
    \textbf{c)} Scaling behavior for the remaining self-supervised models (MAE and Mugs). These models are included here for completeness; all other SSL models’ scaling behaviors are presented in the main paper.}
    \label{fig:task2_position_bias_large_artifacts_base}
\end{figure*}

We extract key embeddings from the Large model variants of MAE, DINOv2, DINOv2+reg, DINOv3, and Mugs to perform unsupervised semantic segmentation (see \cref{fig:task2_position_bias_large_artifacts_base}(a)). Positional bias remains pronounced across all MIM models, while Mugs, a purely discriminative SSL model, serves as a reference baseline exhibiting comparatively little positional bias.
\Cref{fig:task2_position_bias_large_artifacts_base}(b) presents token embeddings from Base model variants, where segmentation artifacts are visible in both CLIP and DINOv2 results, consistent with findings reported in prior work~\cite{yang2024denoising}.
Finally, \Cref{fig:task2_position_bias_large_artifacts_base}(c) provides the remaining SSL models that also lack a clear scaling benefit for unsupervised semantic segmentation, complementing the examples already shown in the main paper. Note that for MAE, the comparison is based on value embeddings from the Base and Large variants, whereas for Mugs, token embeddings from the Base and Large variants are used.



\newpage
\subsection{Unsupervised oversegmentation / -clustering results}
\label{app:unsup_over_seg}
\begin{figure*}[htbp]
    \centering
    \includegraphics[width=1.0\columnwidth]{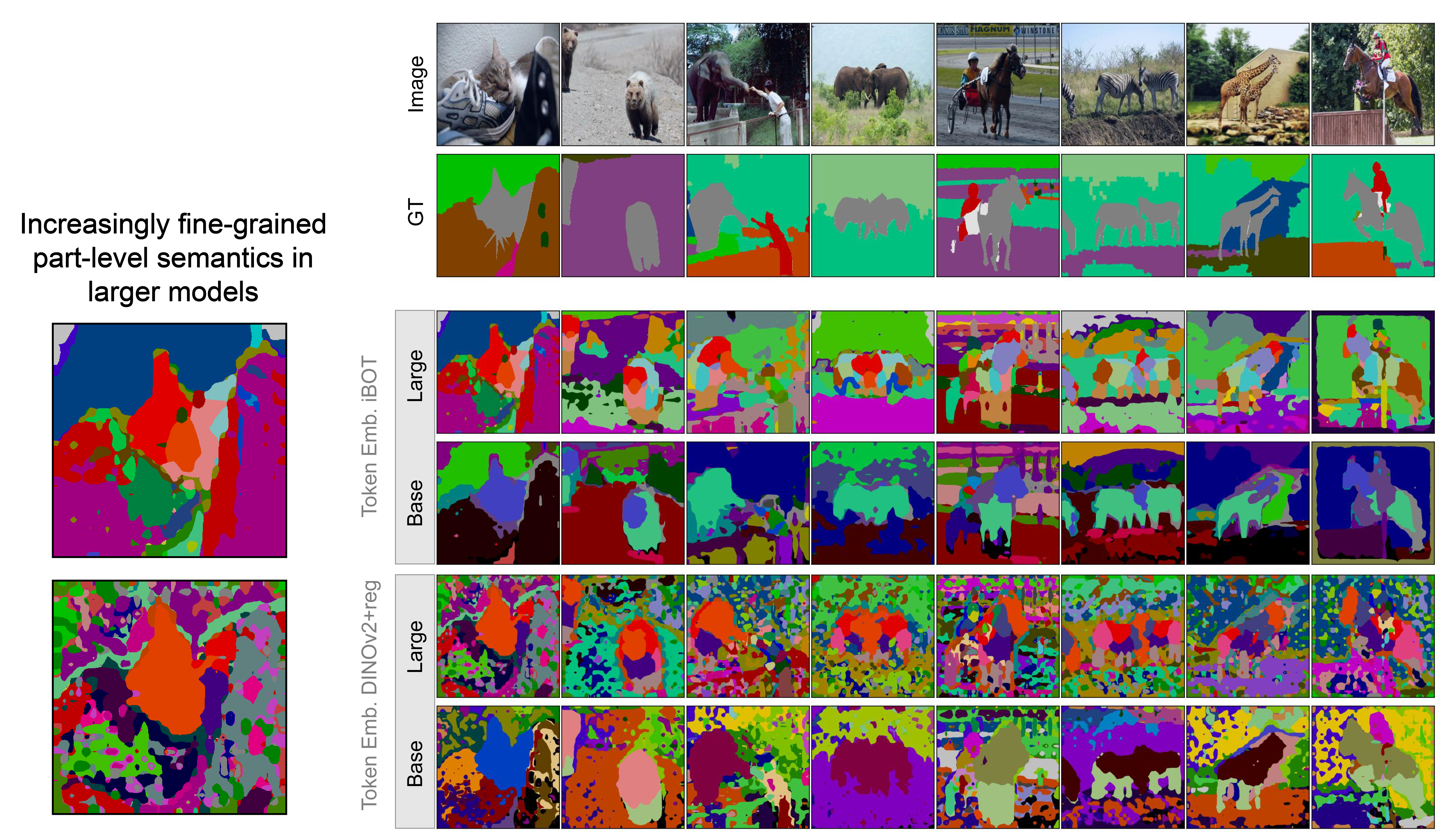}
    \caption{Overclustering of dataset semantics into 97 clusters. Large variants of iBOT and DINOv2+reg exhibit finer-grained part-level clustering, separating regions such as eyes, ears, neck and trunk. In contrast, increasing the number of clusters does not lead to comparable part-level delineation in the Base variants.}
    \label{fig:overseg_animal_samples}
\end{figure*}

We increase the number of clusters by 70, resulting in a total of 97 clusters, in order to induce an overclustering regime on the dataset. This analysis is conducted using the best-layer token embeddings from both the Base and Large variants of Mugs, iBOT, DINOv2, DINOv2+reg, and DINOv3. For brevity, we report only the most salient findings here, focusing on iBOT and DINOv2+reg; see \cref{fig:overseg_animal_samples}.

We find that the Large variants of iBOT and DINOv2+reg exhibit a stronger tendency to delineate fine-grained part-level semantics. For instance, iBOT large model separates regions corresponding to eyes, nose, ears, and neck, while the large model of DINOv2+reg isolates facial and neck regions. Such part-level specialization is substantially less pronounced in the corresponding Base variants.

Furthermore, we observe a qualitative difference in the clustering hierarchy for DINOv2+reg across model scales. The Base variant tends to organize embeddings primarily according to object category, forming clusters that separate animals such as cats, bears, elephants, and horses. In contrast, the Large variant prioritizes part-level semantics, grouping similar anatomical regions (e.g., faces, necks, and legs) across different animal categories, thereby assigning faces from multiple species to the same cluster.

\subsection{Unsupervised semantic segmentation on further datasets}
\label{app:further_datasets}

\begin{figure}[!h]%
    \centering
    \includegraphics[width=\textwidth]{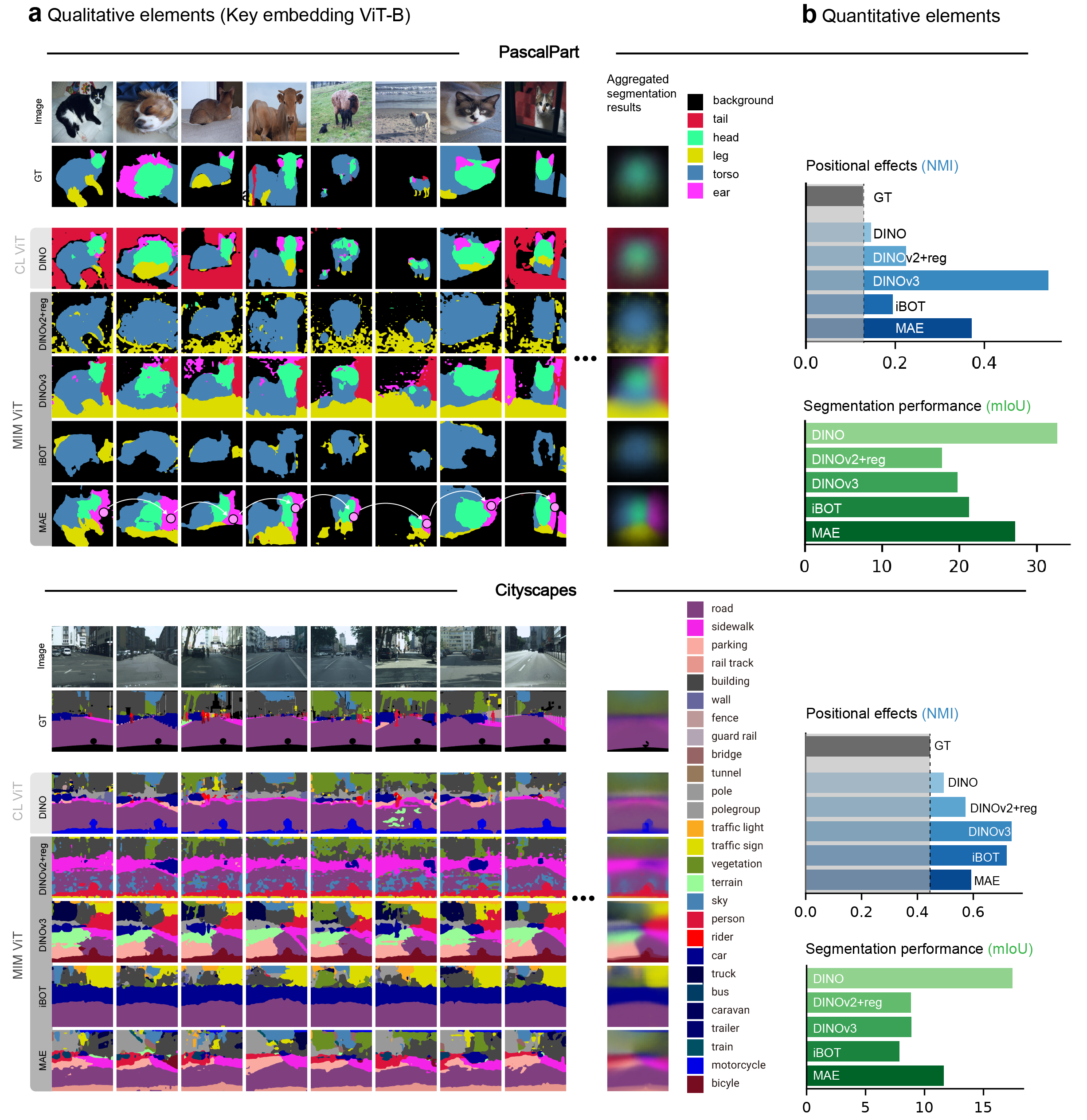} %
    \caption{\textbf{a)} Qualitative examples from the PascalPart and Cityscapes datasets with aggregated maps.
    \textbf{b)} Quantitative results, including positional effects (NMI) and segmentation performance (mIoU) for key embeddings extracted from the best-performing layer of several models on the PascalPart and Cityscapes datasets.}%
    \label{fig:good_datasets}%
\end{figure}

\Cref{fig:good_datasets} replicates the cross-model comparison of key embeddings on the PascalPart ($K=6$) and Cityscapes ($K=27$) datasets using the same experimental protocol as in \cref{suppl:subsec:unsup_sem_seg_training}. These benchmark datasets are also established for unsupervised semantic segmentation, yet have considerably more inherent spatial bias than COCO-Stuff, making them less suitable than COCO-Stuff for model understanding under our protocol.

For PascalPart, we use the animal subset, resulting in approximately 1,077 training images and 1,094 test images.
For both datasets, we adopt the same unsupervised segmentation training settings described in \cref{suppl:subsec:unsup_sem_seg_training}. Specifically, we use a batch size of 24 and initialize the prototypes using patch embeddings sampled from the first 96 mini-batches. Since PascalPart contains fewer than 2,300 images, the prototype initialization effectively uses patch embeddings from the entire training set. We perform a grid search to determine the optimal learning rate and the selected learning rate is $2 \times 10^{-4}$ for both PascalPart and Cityscapes.

\subsection{Zero-shot semantic correspondence setup}
\label{app:semantic_correspondence}

Beyond benchmarking embeddings on unsupervised semantic segmentation task, we take a further step and evaluate them on zero-shot semantic correspondence.
Unlike clustering, which assesses whether embeddings belonging to distinct object categories are globally separable, semantic correspondence requires the model to establish precise, fine-grained spatial matches between semantically related regions across different object instances. This task therefore provides a complementary perspective on embedding quality that reveals whether the learned representations capture not only category-level discriminability but also locally consistent geometric and semantic structure.

\input{tabs/camera_ready_version/spair_pck_per_point}
For each pixel $p_r$ in the reference image, we identify the most similar pixel  $p_t$  in the target image using cosine similarity. To obtain pixel-level representations, patch-wise embeddings of both the reference and target images are bilinearly upsampled to pixel resolution.  The corresponding pixel $p_t$ is predicted as
\begin{equation}
    \text{match}(p_r) = \arg\max_{p_t} \; 
 \frac{  \hat{\mathbf{f}}_i^{(p_r)}   \cdot\hat{\mathbf{f}}_j^{(p_t)} }{\left\|\hat{\mathbf{f}}_i^{(p_r)} \right\| \left\|\hat{\mathbf{f}}_j^{(p_t)} \right\|}
\end{equation}
where $(i,j)$ denotes a reference-target image pair and $\hat{\mathbf{f}}_i^{(p_r)} \in \mathbb{R}^{d}$ denotes the feature vector at 
pixel $p_r$. 
Layer-wise performance of all models is shown in~\cref{fig:task1_kqvt} in~\cref{app:task1_layerwise}. We summarize the best-layer performance for each type of embedding in ~\cref{tab:spair_perpoint}. Clear scaling behavior can be observed, which was not reflected in our unsupervised semantic segmentation task (see~\cref{fig:fig_performance_task_2}). 
Qualitative analysis on results between the Base and Large variants are provided in~\cref{app:task1_scaling}.

\subsection{Zero-shot semantic correspondence layer-wise performance}
\label{app:task1_layerwise}

\begin{figure}[h]%
    \centering
    \subfloat[\centering key embedding]{{\includegraphics[width=6.4cm]{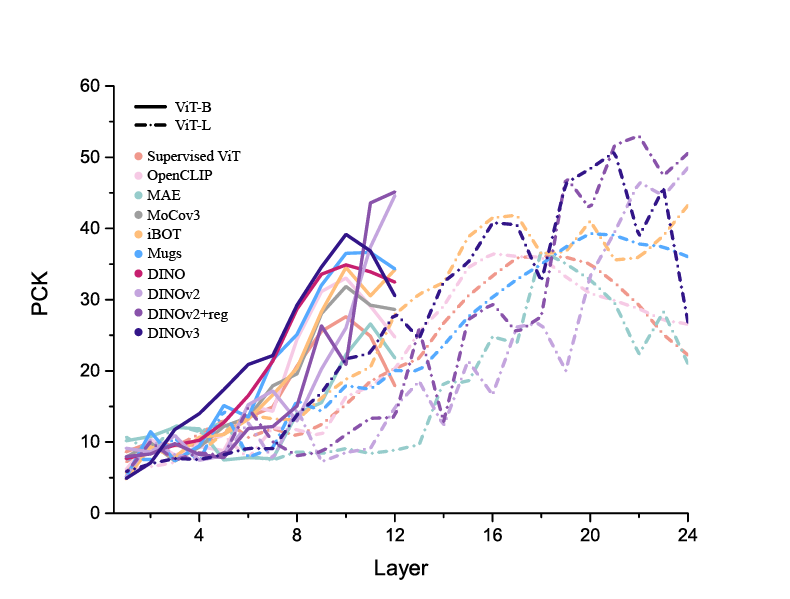} }}%
    \subfloat[\centering query embedding]{{\includegraphics[width=6.4cm]{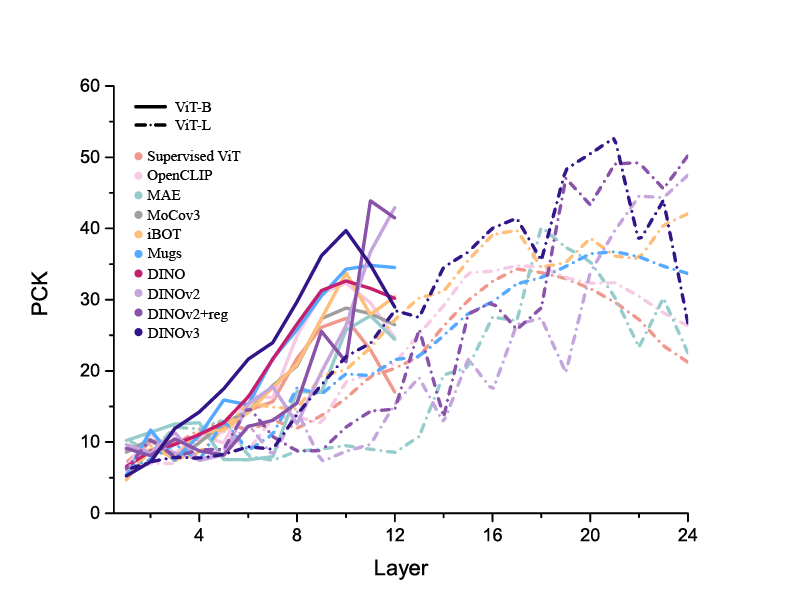} }}%
    \\
    \subfloat[\centering value embedding]{{\includegraphics[width=6.4cm]{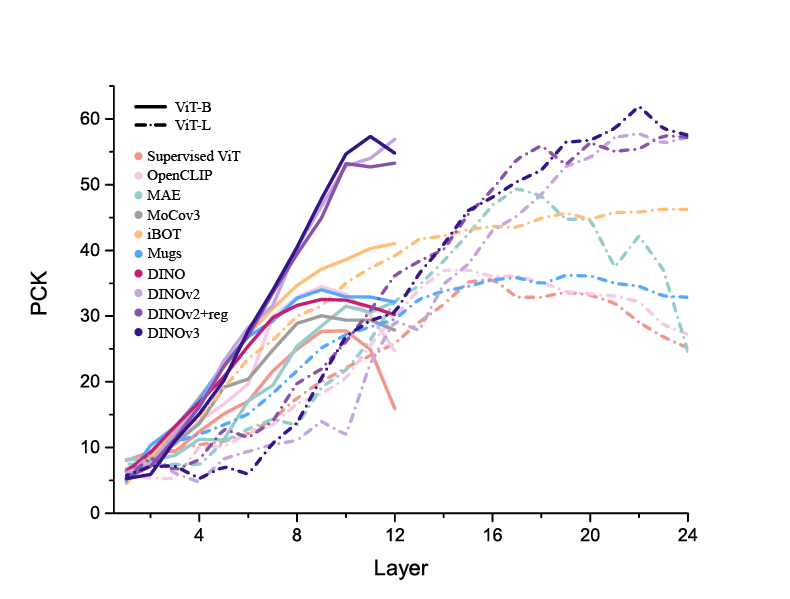} }}%
    \subfloat[\centering token embedding]{{\includegraphics[width=6.4cm]{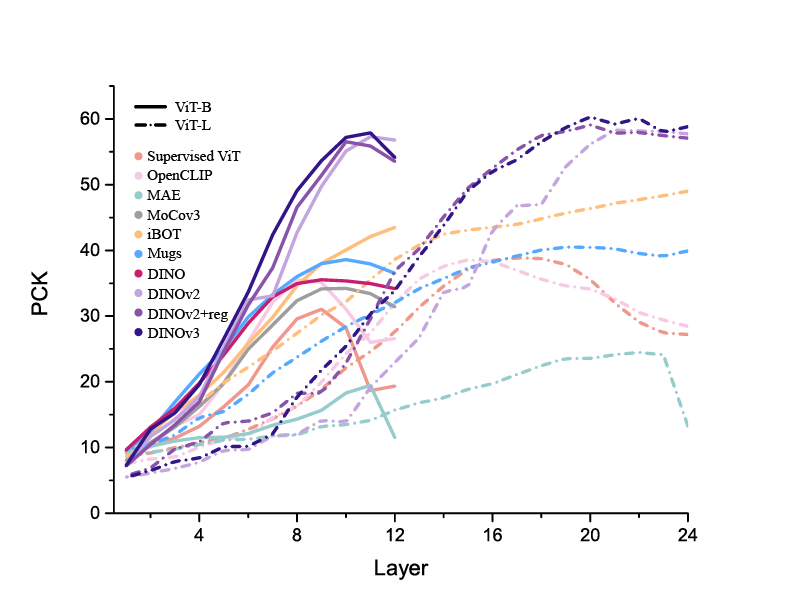} }}%
    \caption{Performance of \textbf{key}, \textbf{query}, \textbf{value} and \textbf{token} embeddings on \textbf{zero-shot semantic correspondence} task across layers and model sizes.}%
    \label{fig:task1_kqvt}%
\end{figure}

\Cref{fig:task1_kqvt} presents the layer-wise performance on zero-shot semantic correspondence for both base and large models across all four embedding types. We observe positive scaling behavior in this task, where the best-performing layer of the large model consistently outperforms that of the base model for the same embedding type.

\subsection{Final-layer degradation in zero-shot semantic correspondence: Supervised vs. SSL Models}
\label{app:final_layer_degradation_task1}

\begin{figure*}[h!]
\centering
\includegraphics[width=1.0\columnwidth]{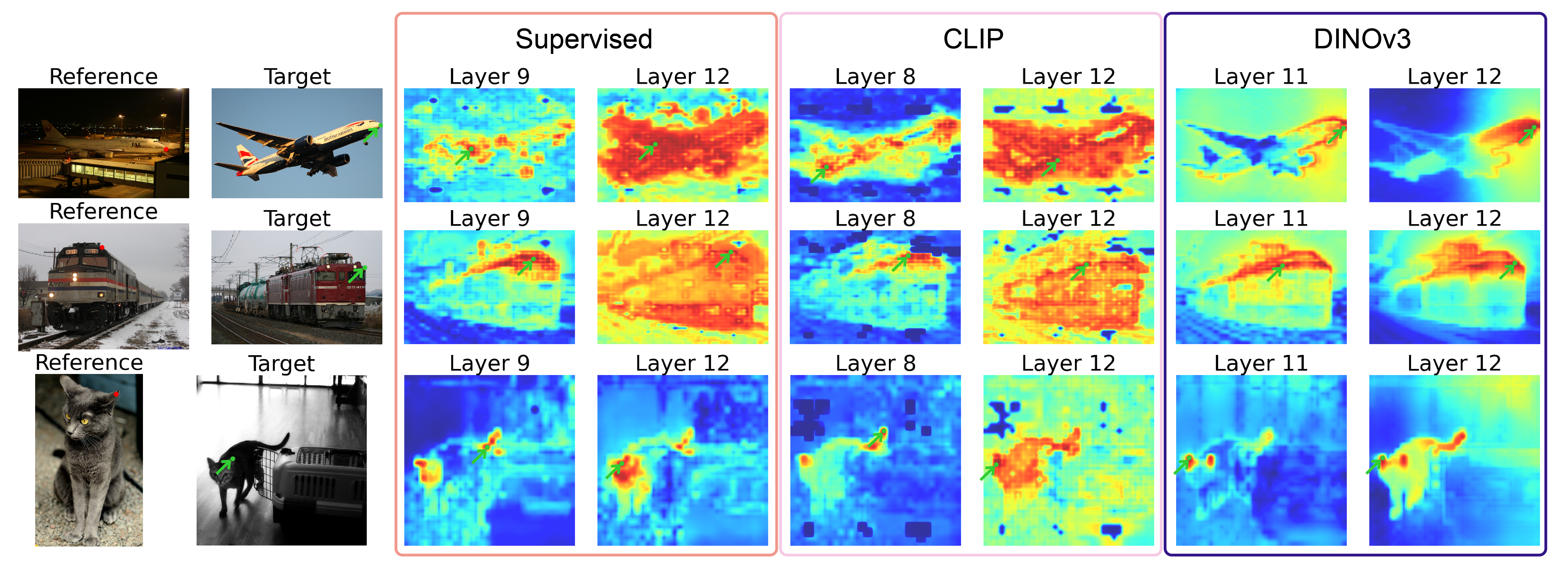}
\caption{Semantic correspondence qualitative results of \textbf{token} embeddings from three representative SSL ViT base models. We present three example image pairs to compare the qualitative performance of the \textbf{best-performing} layer and the \textbf{final} layer embeddings from (a) Supervised ViT, (b) CLIP, and (c) DINOv3. Heatmaps visualize the cosine similarity between a selected keypoint in the reference image (\textcolor{red}{red} dot) and all spatial locations in the target image. The ground-truth matching point in the target image and the highest-reacted points in heatmaps are marked in \textcolor{green}{green}.
\label{fig:task1_1_qual}}
\end{figure*}

~\Cref{tab:spair_perpoint} shows \textbf{token} embeddings achieve the best performance across all embedding types in the vast majority of models, with the exception of MAE and DINOv2-Large. Consistent with the findings in~\cref{fig:fig_performance_task_2}, the best-performing layer for zero-shot semantic correspondence rarely coincides with the final layer. Nevertheless, most SSL models exhibit only a mild performance decline after the peak—with iBOT even continuing to improve. In contrast, supervised ViT and CLIP suffer a more substantial drop beyond their optimal layers.

To better understand this divergence, we conduct a qualitative analysis using three image pairs, visualizing activation heatmaps from the best-performing and final layers (see \cref{fig:task1_1_qual}). We include the top-performing model, DINOv3, together with the two models exhibiting the strongest degradation: supervised ViT and CLIP. Our analysis reveals that DINOv3 embeddings maintain relatively fine-grained, part-aware representations from the intermediate best layer to the final layer. In contrast, supervised ViT and CLIP produce final-layer token embeddings that are often over-activated across spatial locations, including irrelevant background regions.

Given that supervised ViT is trained with image‑class labels and CLIP with an image–text contrastive objective, we hypothesize that their final-layer embeddings—being closer to the output head—tend to encode high‑level semantic summaries, frequently at the expense of fine‑grained spatial and part‑level detail. This is likely a consequence of optimizing for global image–text/label alignment rather than for local patch‑augmentation invariance, as in a discriminative SSL objective.



\subsection{Zero-shot semantic correspondence scaling behavior
\label{app:task1_scaling}}

\begin{figure*}[htbp]
    \centering
    \includegraphics[width=\columnwidth]{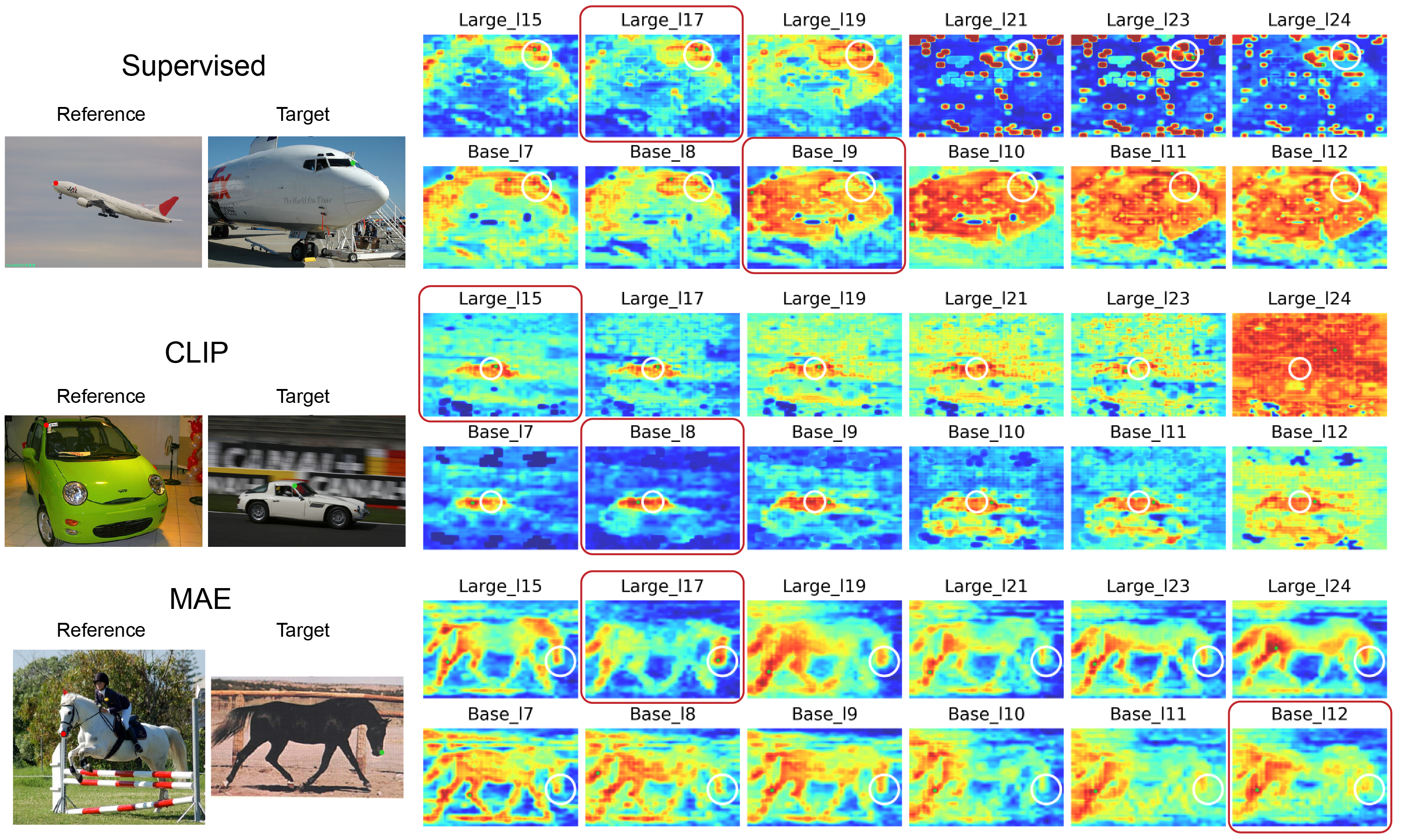}
    \caption{Semantic correspondence results using supervised ViT (token), CLIP (token), and MAE (value) features. \textcolor{green}{green} marks the ground-truth correspondence in the target image and the highest-response locations in the heatmaps. White circles indicate regions counted as true positives under the PCK metric. The best-performing layer is highlighted with a  \textcolor{red}{red} rectangle. In the Large variants, supervised ViT feature correspondence heatmaps in layers 21, 23, and 24 exhibit multiple highly activated block-like regions that are semantically irrelevant, while the CLIP-Large heatmap in the final layer shows overactivation across the entire image region. In contrast, their Base variants produce more robust and meaningful responses, selectively highlighting the main objects and semantically relevant regions across the last several layers.}
    \label{fig:task1_bl_compare_supervised_clip_mae}
\end{figure*}

\begin{figure*}[htbp]
    \centering
    \includegraphics[width=\columnwidth]{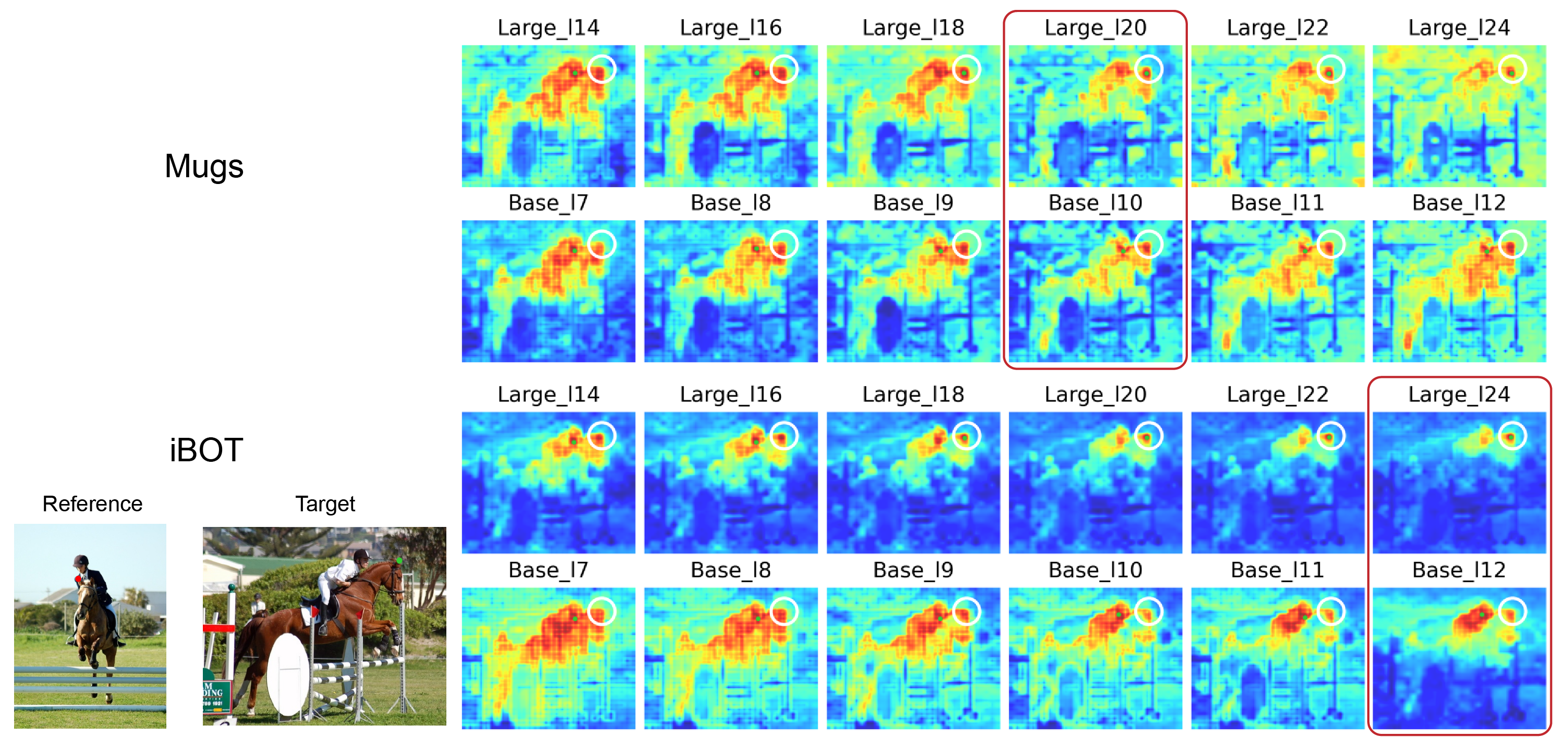}
    \caption{Semantic correspondence results using Mugs (token) and iBOT (token) features. The iBOT large model captures finer semantic structures compared to the base model. Same notation as \cref{fig:task1_bl_compare_supervised_clip_mae}.}
    \label{fig:task1_bl_compare_mugs_bot}
\end{figure*}

\begin{figure*}[htbp]
    \centering
    \includegraphics[width=\columnwidth]{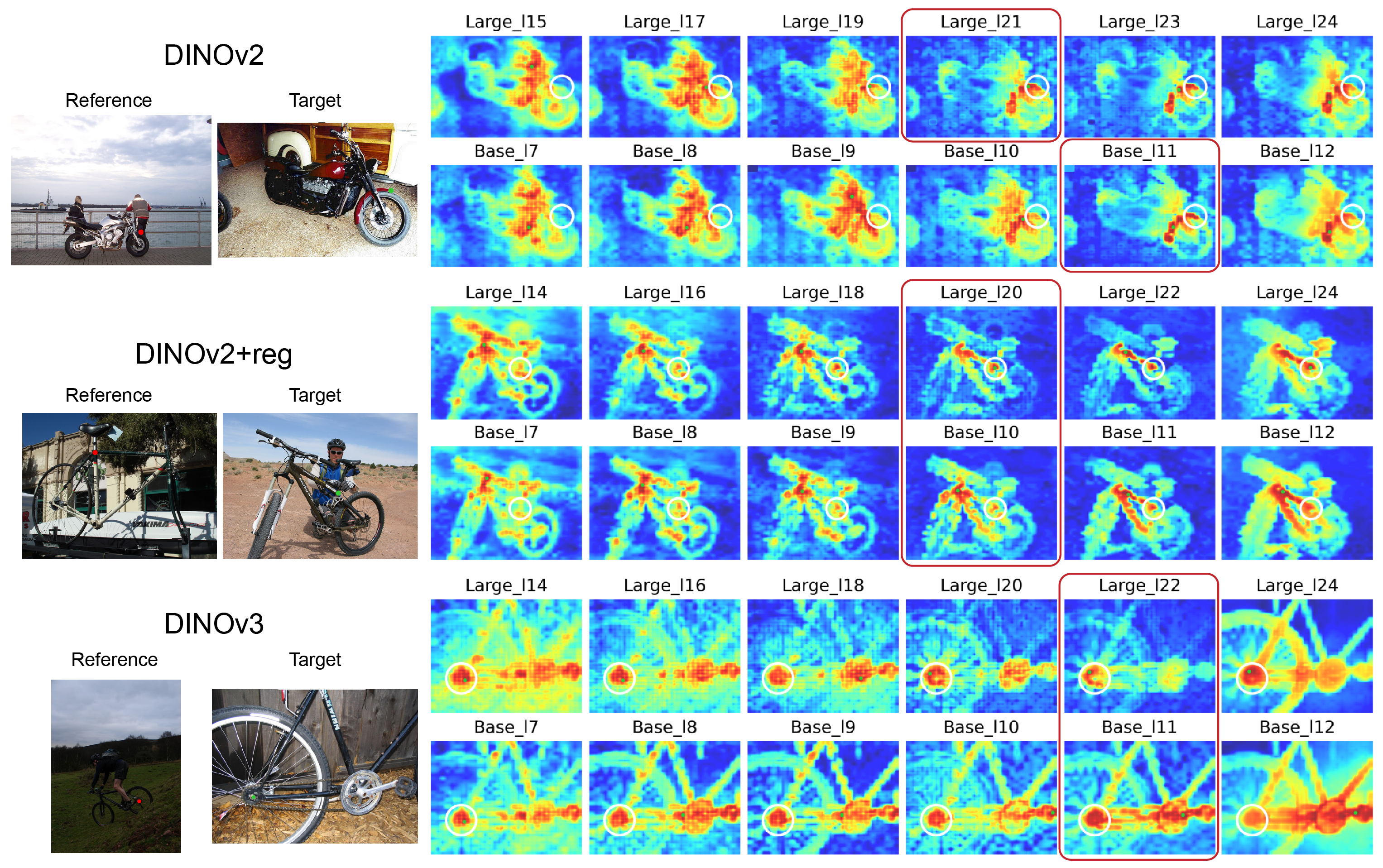}
    \caption{Semantic correspondence results using DINOv2 (token),  DINOv2+reg (token) and DINOv3 (token) features. Same notation as \cref{fig:task1_bl_compare_supervised_clip_mae}.}
    \label{fig:task1_bl_compare_dino_family}
\end{figure*}

We analyze cases where the Large variants outperform the corresponding Base variants when using their best-performing feature layers for the zero-shot semantic correspondence task. For supervised models (Supervised ViT and CLIP), although the Large variants achieve higher performance, primarily due to more localized high-response correspondence points, we observe that the last several layers of the large models exhibit substantially stronger artifacts. These artifacts often manifest as over-activation across the entire image or as randomly distributed high-response regions, as shown in~\cref{fig:task1_bl_compare_supervised_clip_mae}.

In contrast, in \cref{fig:task1_bl_compare_mugs_bot}, the iBOT Large variant produces more localized and accurate responses. This suggests that the model captures finer structural details, which is consistent with our observations from the unsupervised segmentation and overclustering analyses.
\cref{fig:task1_bl_compare_mugs_bot} also illustrates a representative failure case, where the reference point lies within a patch containing both human and horse ear semantics; incorrect predictions frequently yield high responses in the human chest or shoulder rather than the horse ear. This error stems partly from patch-based tokenization in Vision Transformers: since each patch is encoded as a single embedding, patches spanning multiple semantic regions produce mixed representations. Bilinear upsampling yields dense feature maps but does not restore the original semantic resolution, as pixel-level features may still reflect information aggregated over the entire patch, making correspondence ambiguous when reference points fall within semantically mixed patches.

For completeness, we also include qualitative results for DINOv2, DINOv2+reg, and DINOv3, where the Large variants are generally better, owing to slightly more focused, fine-grained semantic activation(See \cref{fig:task1_bl_compare_dino_family}).


\subsection{Effect of Layer Normalization on final ViT token embeddings}
\label{app:layernorm_effect}
We analyze the effect of the final Layer Normalization (LayerNorm) on the output token embeddings of Vision Transformers (ViTs) across two tasks: unsupervised semantic segmentation (see~\cref{tab:effect_layernorm_task2}) and zero-shot semantic correspondence (see~\cref{tab:effect_layernorm_task1}). Specifically, we compare feature representations extracted from the final transformer block \emph{before} and \emph{after} the LayerNorm operation.

Our results show that LayerNorm has a pronounced impact on unsupervised semantic segmentation performance, most notably for DINOv3. In contrast, its effect on zero-shot semantic correspondence is comparatively modest. We leave a deeper investigation into the underlying mechanism and in particular why DINOv3 exhibits such sensitivity as an avenue for future work.
\input{tabs/camera_ready_version/effect_layernorm_task2}
\input{tabs/camera_ready_version/effect_layernorm_task1}

%% file: tabs/camera_ready_version/ssl_table_1.tex
\setlength{\tabcolsep}{12pt} 
\begin{table*}
\centering
\caption{Summary of selected self-supervised and supervised learning models with their basic information such as available model sizes, pre-training dataset, positional embedding types and  training paradigm category.
 \label{tab:ssl_models_tab1}
}

    \resizebox{1\linewidth}{!}{%
    \begin{tabular}{ lllll} 
    \noalign{\hrule height 2pt}
     SSL framework& Size & Dataset & Pos. Emb. & Category  \\ 
    \midrule

    Supervised ViT~\cite{dosovitskiy2020image} &  B,L & ImageNet-21K & Learned abs & Supervised\\
    
    OpenCLIP ~\cite{ilharco2021openclip}   &B,L,H,G & LAION-2B& Learned abs & Supervised \\

    MaskAutoEncoder ~\cite{he2022masked} &B,L,H & ImageNet-1K & Sin-Cos&  MIM\\

    MoCov3 ~\cite{chen2021empirical}    &B& ImageNet-1K&  Sin-Cos & CL \\ 

    Mugs ~\cite{zhou2022mugs}   &B,L & ImageNet-1K& Learned abs & CL \\ 

    iBOT ~\cite{zhou2021ibot} & B,L & ImageNet-1K & Learned abs& MIM \\ 
    
    DINO ~\cite{caron2021emerging}   &B  &ImageNet-1K & Learned abs & CL \\ 
    
    DINOv2 ~\cite{oquab2023dinov2}    &B,L,G & LVD-142M & Learned abs& MIM \\ 

    DINOv2+reg ~\cite{darcet2023vision} & B,L,G & LVD-142M& Learned abs &  MIM \\

    DINOv3 ~\cite{simeoni2025dinov3} & B,L,H,7B& LVD-1689M& Axial RoPE	& MIM  \\
    \bottomrule
    \end{tabular}
    }

\end{table*}

%% file: tabs/camera_ready_version/ssl_table_2.tex
\begin{table}
\centering
\caption{Selected self-supervised and supervised learning models and their distinctive properties.
\label{tab:ssl_models_tab2}}
\fontsize{7}{8}\selectfont
\setlength{\tabcolsep}{4pt}
\renewcommand{\arraystretch}{1.05}

\begin{tabularx}{\linewidth}{lX}
\toprule
\textbf{SSL framework} & \textbf{Unique property} \\
\midrule

Supervised ViT ~\cite{dosovitskiy2020image} &
Supervised classification training \\

OpenCLIP ~\cite{ilharco2021openclip} &
Joint image-text representation learning on large-scale image-text pairs \\

MAE ~\cite{he2022masked} &
Reconstruct missing parts of an image from masked input patches \\

MoCov3 ~\cite{chen2021empirical} &
Contrast different augmented views of the image in a batch using momentum-updated encoder \\

Mugs ~\cite{zhou2022mugs} &
Explicitly learn multi-granular visual features through complementary supervision \\

iBOT  ~\cite{zhou2021ibot} &
Uses masked image modeling and self-distillation to learn strong visual representations without labels \\

DINO ~\cite{caron2021emerging} &
Self-distillation with no labels using a teacher-student setup and contrastive learning \\

DINOv2 ~\cite{oquab2023dinov2}  &
Scales up DINO with larger curated data and adds masked image modeling for better general-purpose features \\

DINOv2+reg ~\cite{darcet2023vision} &
Extends DINOv2 by introducing learnable register tokens to eliminate attention map artifacts and capture global context \\

DINOv3 ~\cite{simeoni2025dinov3} &
Improves upon DINOv2 by scaling up the model and dataset and introducing “Gram anchoring” to get smoother features \\

\bottomrule
\end{tabularx}
\end{table}

%% file: tabs/camera_ready_version/spair_pck_per_point.tex
\begin{table}[h]
\centering
\caption{
PCK($\alpha_{bbox}=0.1$) \texttt{per} \texttt{point} on SPair-71k. We report, for each model, the maximum PCK obtained across layers for each embedding type. \textbf{Best layer} denotes the depth at which the model achieves its overall best matching performance on its best performing embedding. For ViT-L, $\delta$ indicates the performance gain compared to the corresponding ViT-B model (same method).
}
\setlength{\tabcolsep}{3pt}
\resizebox{1\linewidth}{!}{%
\begin{tabular}{ clcccc|c } 
\toprule 
\multirow{2}{*}{\shortstack[c]{\textbf{Model Size}}} & \multirow{2}{*}{\textbf{Method}} & \multicolumn{4}{c}{\textbf{Facet Category}} & \multirow{2}{*}{\textbf{Best layer}} \\ 
\cmidrule(lr){3-6}
& & Key & Query & Value & Token &  \\  
\midrule 

\multirow{10}{*}{ViT-B} & 
Supervised ViT~\cite{dosovitskiy2020image} & 27.62 & 27.75 & 27.36 & \best{31.03} & 9/12 \\

& OpenCLIP~\cite{ilharco2021openclip} & 33.05 & 32.34 & 34.46 & \best{35.57} & 8/12 \\

& MaskAutoEncoder~\cite{he2022masked} & 26.60 & 27.73 & \best{32.16} & 19.48 & 12/12 \\

& MoCov3~\cite{chen2021empirical} & 31.85 & 28.82 & 30.05 & \best{34.22} & 10/12 \\

& Mugs~\cite{zhou2022mugs} & 36.63 & 34.81 & 34.01 & \best{38.61} & 10/12 \\

& DINO~\cite{caron2021emerging} & 34.89 & 32.64 & 32.54 & \best{35.54} & 9/12 \\

& iBOT~\cite{zhou2021ibot} & 34.52 & 33.76 & 41.03 & \best{43.50} & 12/12 \\

& DINOv2~\cite{oquab2023dinov2} & 44.63 & 42.99 & 56.92 & \best{57.28} & 11/12 \\

& DINOv2+reg ~\cite{darcet2023vision} & 45.14 & 43.90 & 53.28 & \best{56.52} & 10/12 \\

& DINOv3~\cite{simeoni2025dinov3} & 39.16 & 39.71 & 57.34 & \best{57.87} & 11/12 \\

\midrule 

\multirow{8}{*}{ViT-L} & 
Supervised ViT~\cite{dosovitskiy2020image} 
& 36.12~\posd{+8.50} 
& 34.30~\posd{+6.55} 
& 35.66~\posd{+8.30} 
& \best{38.84}~\posd{+7.81} 
& 17/24 \\

& OpenCLIP~\cite{ilharco2021openclip} 
& 36.48~\posd{+3.43} 
& 34.76~\posd{+2.42} 
& 36.96~\posd{+2.50} 
& \best{38.55}~\posd{+2.98} 
& 15/24 \\

& MaskAutoEncoder~\cite{he2022masked} 
& 36.83~\posd{+10.23} 
& 40.23~\posd{+12.50} 
& \best{49.41}~\posd{+17.25} 
& 24.45~\posd{+4.97} 
& 17/24 \\

& Mugs~\cite{zhou2022mugs} 
& 39.30~\posd{+2.67} 
& 36.73~\posd{+1.92} 
& 36.19~\posd{+2.18} 
& \best{40.46}~\posd{+1.85} 
& 20/24 \\

& iBOT~\cite{zhou2021ibot} 
& 43.26~\posd{+8.74} 
& 42.08~\posd{+8.32} 
& 46.27~\posd{+5.24} 
& \best{48.99}~\posd{+5.49} 
& 24/24 \\

& DINOv2~\cite{oquab2023dinov2} 
& 48.55~\posd{+3.92}
& 47.51~\posd{+4.52} 
& 57.76~\posd{+0.84} 
& \best{58.28}~\posd{+1.00} 
& 21/24 \\

& DINOv2+reg~\cite{darcet2023vision} 
& 53.01~\posd{+7.87} 
& 50.23~\posd{+6.33} 
& 57.47~\posd{+4.19} 
& \best{59.09}~\posd{+2.57} 
& 20/24 \\

& DINOv3~\cite{simeoni2025dinov3} 
& 50.76~\posd{+11.60}
& 52.74~\posd{+13.03}
& \best{61.95}~\posd{+4.61} 
& 60.28~\posd{+2.41} 
& 22/24 \\

\bottomrule
\end{tabular}
}
\label{tab:spair_perpoint}
\end{table}

%% file: tabs/camera_ready_version/effect_layernorm_task2.tex
\begin{table}[h]
\centering
\setlength{\tabcolsep}{2pt}
\caption{Effect of Layer Normalization on final ViT \textbf{token} embeddings for unsupervised semantic segmentation.}
\resizebox{1\linewidth}{!}{%
\begin{tabular}{c|cccccccccccccccccc}
\toprule
 & \multicolumn{2}{c}{\textbf{Supervised}} 
 & \multicolumn{2}{c}{\textbf{CLIP}} 
 & \multicolumn{2}{c}{\textbf{MAE}} 
 & \multicolumn{1}{c}{\textbf{MoCov3}} 
 & \multicolumn{2}{c}{\textbf{iBOT}} 
 & \multicolumn{2}{c}{\textbf{Mugs}} 
 & \multicolumn{1}{c}{\textbf{DINO}} 
 & \multicolumn{2}{c}{\textbf{DINOv2}} 
 & \multicolumn{2}{c}{\textbf{DINOv2+reg}} 
 & \multicolumn{2}{c}{\textbf{DINOv3}} \\
\cmidrule(lr){2-3}
\cmidrule(lr){4-5}
\cmidrule(lr){6-7}
\cmidrule(lr){8-8}
\cmidrule(lr){9-10}
\cmidrule(lr){11-12}
\cmidrule(lr){13-13}
\cmidrule(lr){14-15}
\cmidrule(lr){16-17}
\cmidrule(lr){18-19}
 & Base & Large  
 & Base & Large 
 & Base & Large 
 & Base
 & Base & Large 
 & Base & Large 
 & Base  
 & Base & Large 
 & Base & Large 
 & Base & Large  \\

\midrule
Before  
 & 9.09 & 9.95 
 & 15.04 & 10.78 
 & 2.59 & 0.85 
 & 17.84
 & 16.42 & 15.89 
 & 12.77 & 4.99 
 & 15.04  
 & 21.02 & 18.94 
 & 18.11 & 15.8 
 & 1.45 & 3.84 \\
 
After
 & 12.16 & 11.7 
 & 18.52 & 9.41 
 & 5.14 & 5.54 
 & 11.46
 & 14.05 & 13.73 
 & 13.84 & 10.85 
 & 16.11 
 & 19.71 & 17.9 
 & 20.15 & 18.24 
 & 16.79 & 19.59 \\
\midrule
Diff.
 & \posc{+3.07} & \posc{+1.75}
 & \posc{+3.48} & \negc{-1.37}
 & \posc{+2.55} & \posc{+4.69}
 & \negc{-6.38}
 & \negc{-2.37} & \negc{-2.16}
 & \posc{+1.07} & \posc{+5.86}
 & \posc{+1.07}
 & \negc{-1.31} & \negc{-1.04}
 & \posc{+2.04} & \posc{+2.44}
 & \posc{+15.34} & \posc{+15.75} \\
\bottomrule

\end{tabular}
}
\label{tab:effect_layernorm_task2}
\end{table}

%% file: tabs/camera_ready_version/effect_layernorm_task1.tex
\begin{table}[h]
\centering
\setlength{\tabcolsep}{2pt}
\caption{Effect of Layer Normalization on final ViT \textbf{token} embeddings for zero-shot semantic correspondence.}
\resizebox{1\linewidth}{!}{%
\begin{tabular}{c|cccccccccccccccccc}
\toprule
 & \multicolumn{2}{c}{\textbf{Supervised}} 
 & \multicolumn{2}{c}{\textbf{CLIP}} 
 & \multicolumn{2}{c}{\textbf{MAE}} 
 & \multicolumn{1}{c}{\textbf{MoCov3}} 
 & \multicolumn{2}{c}{\textbf{iBOT}} 
 & \multicolumn{2}{c}{\textbf{Mugs}} 
 & \multicolumn{1}{c}{\textbf{DINO}} 
 & \multicolumn{2}{c}{\textbf{DINOv2}} 
 & \multicolumn{2}{c}{\textbf{DINOv2+reg}} 
 & \multicolumn{2}{c}{\textbf{DINOv3}} \\
\cmidrule(lr){2-3}
\cmidrule(lr){4-5}
\cmidrule(lr){6-7}
\cmidrule(lr){8-8}
\cmidrule(lr){9-10}
\cmidrule(lr){11-12}
\cmidrule(lr){13-13}
\cmidrule(lr){14-15}
\cmidrule(lr){16-17}
\cmidrule(lr){18-19}
 & Base & Large  
 & Base & Large 
 & Base & Large 
 & Base
 & Base & Large 
 & Base & Large 
 & Base  
 & Base & Large 
 & Base & Large 
 & Base & Large  \\

\midrule
Before  
 & 17.39 & 26.27 
 & 25.18 & 28.81 
 & 18.78 & 17.82 
 & 32.34
 & 42.65 & 48.79 
 & 35.47 & 38.01 
 & 34.66  
 & 56.94 & 57.11 
 & 53.55 & 55.08 
 & 57.01 & 56.66 \\
 
After
 & 19.34 & 27.19
 & 26.57 & 28.46 
 & 11.5 & 13.22 
 & 31.42
 & 43.5 & 48.99 
 & 36.52 & 39.92 
 & 34.19  
 & 56.8 & 57.74
 & 53.55 & 57.08 
 & 54.12 & 58.84 \\
\midrule
Diff.
 & \posc{+1.95} & \posc{+0.92}
 & \posc{+1.39} & \negc{-0.35}
 & \negc{-7.28} & \negc{-4.60}
 & \negc{-0.92}
 & \posc{+0.85} & \posc{+0.20}
 & \posc{+1.05} & \posc{+1.91}
 & \negc{-0.47}
 & \negc{-0.14} & \posc{+0.63}
 & \zeroc{0.00} & \posc{+2.00}
 & \negc{-2.89} & \posc{+2.18} \\
\bottomrule

\end{tabular}
}
\label{tab:effect_layernorm_task1}
\end{table}